**PREPRINT VERSION 2. THIS VERSION INCLUDES SUBSTANTIAL REVISIONS TO THE THEORETICAL FRAMING, DISCUSSION, AND SUPPLEMENTARY MATERIAL.**

# Movement Information from a Robotic Effector Influences Human Performance During Object Handovers

Róisín Keenan, Joost C. Dessing

School of Psychology, Queen's University Belfast

**ABSTRACT**

Recent advances in robotics have increased opportunities for humans and robots to collaborate during object manipulation tasks. Effective coordination in such interactions depends on how humans use movement information provided by a robotic partner to organize and adjust their own actions. The present study examined how movement characteristics of a robotic effector influence human motor performance during robot-to-human handovers. Using a virtual reality-based handover paradigm, three sets of factors were investigated: (1) alternative modes of spatiotemporal control of object handovers, (2) robot velocity profile characteristics (minimum jerk, constant velocity, constant acceleration, and biphasic), and (3) the timing with which the required pickup orientation became visually specified. Across experiments, performance benefited when task-relevant action requirements were specified earlier in the interaction and when robot movements provided information that could be readily used for coordination. Smooth velocity profiles also improved performance, with benefits that appeared to be concentrated in the final deceleration phase preceding object acquisition rather than reflecting human-like motion more generally. Together, the results highlight the importance of movement information, action-requirement specification, and movement kinematics in supporting coordinated action. These findings contribute to understanding how humans organize and adapt their movements during interactions with robotic systems, while providing principles for the design of more effective human-robot collaborations.

## INTRODUCTION

Object handovers are a common component of collaborative human-robot tasks and require humans to coordinate their movements to those of a robotic effector [33, 64]. Different movement characteristics of robotic effectors may influence human motor performance in different ways [44]. Understanding how and why is important for designing effective and intuitive collaborative robotic systems, which are becoming increasingly relevant in society [5]. The current study examined how the timing and kinematic characteristics of a robotic effector influence human motor performance during robot-to-human object handovers, with a particular focus on how movement information provided by the robot effector supports coordination [6, 11, 49, 54].

When interacting with others, humans or robots, the timing of movement initiation can strongly influence coordination [4]. During handover tasks, the partner initiating movement provides visual information about the future spatiotemporal evolution of the interaction [15, 20]. This information becomes available before the responding partner begins their own movement and may therefore contribute to movement organization and online control. Earlier access to such information may enable more effective timing and positioning of the responding movement. Although leadership structures and initiative distributions have been studied extensively in human-robot interaction [45, 47], fewer studies have directly examined how different initiation modes influence objective motor performance during handovers.

When one partner initiates the interaction, a subsequent question is how available movement information is used to coordinate actions in both space and time. Object handovers can be viewed as a form of manual interception in which the receiving partner must arrive at the interaction location at an appropriate time to acquire the object. Successful interception depends on coordinating one's movement with the motion of the target object [19], suggesting that temporal alignment between interaction partners may facilitate handover performance. Coordinated behaviour in both human-human interactions and HRI is frequently associated with temporal synchronization between interaction partners [41, 46, 61]. Modern robotic systems can adapt their movement online in response to the human partner, allowing synchronization to be actively maintained rather than emerging spontaneously [24, 40, 64]. Such adaptive synchronization may be particularly beneficial during handovers, where the timing of object arrival directly affects the difficulty of acquisition [41]. This highlights the potential importance of temporal alignment in facilitating smooth and coordinated handovers, which may generalize to interactions with robotic effectors. This motivates an explicit comparison of different forms of temporal alignment.

The exact movement of a robotic effector may influence human performance both through the available visual information to the human partner – the visual movement pattern – and control preferences (closely related to the aforementioned synchronization). Movement kinematics provide rich cues about goals, intentions, and internal states [7, 28, 55, 60], which humans can readily extract from biological motion [7, 35] to support smooth interpersonal interactions with humans and robots [54, 56]. Glasauer et al. (2010) found that performance (i.e., end-effector positioning) in a human-robot handover task benefited from robotic movement having a human-like smooth (i.e., minimal jerk) velocity profile by reducing reaction times compared to conventional trapezoidal velocity profiles [29]. This could indicate that familiar, smooth robotic movement kinematics benefit interactions with robotic effectors, possibly because such movements are easier to interpret and coordinate with [15, 38, 56, 60].

Viewing handover tasks as a form of manual interception predicts that the end-effector or object velocity profile will influence the movement [9, 10, 21], especially when the exact interaction location is undefined [16]. Indeed, successful interception – occurring at the right place at the right time – requires accounting for how the robotic effector moved the object as a function of time [9, 19, 53]. Research on interception has examined how target velocity affects performance, often using constant velocity profiles as fully predictable motion patterns [21, 62]. Observers generally perform better when tracking and intercepting objects moving at constant velocities compared to those with accelerating motion [25], except for gravitational acceleration [39, 57]. Importantly, neither real-world objects nor biological entities move with constant velocity and the arguments above might suggest handover tasks would benefit from smooth velocity profiles with a gradual increase and subsequent gradual decrease. Research on manual interception suggests this benefit might related to the smooth final deceleration, which would afford precise positioning when the to-be-picked-up object shows smaller displacement [25, 52]. These arguments call for a detailed comparison of handover performance for different robot effector velocity profiles.

The final factor considered here relates to the information contained within partner movements and the timing with which task-relevant action requirements become specified [1, 63]. As argued above, earlier specification of such information is likely to facilitate task performance. In the context of humans interacting with robotic effectors, the distinction between legible and predictable motion has been made [20]. *Predictability* pertains to the probable movement given our knowledge of the goal, while *legibility* pertains to the probable goal or action requirements given a partially observed movement of the robotic effector. The central assumption is that earlier legibility of one partner's movements facilitates the performance of the other partner. For instance, when picking up objects in different orientations, humans would likely benefit when the required pickup orientation (specified by the final object orientation) becomes apparent earlier during the movement [20]. This motivates an investigation of how the timing with which the required pickup orientation becomes visually specified influences human motor performance during the robot-to-human handover task.

The present study thus examined how different movement characteristics of a robotic effector influence human motor performance during object handovers. Specifically, we focused on a robot-to-human handover task in which participants acquired a small object from a robotic effector and deposited it into a receptacle. To systematically manipulate movement characteristics while maintaining experimental control, all experiments were conducted in a virtual reality (VR) environment. Although the present study was exploratory in nature, previous work on movement coordination and manual interception suggests that movement characteristics providing earlier, more salient, and more readily interpretable movement information may be associated with improved human motor performance during object handovers. Specifically, we examined the effects of (1) alternative modes of spatiotemporal control during object handovers, (2) velocity profile characteristics, and (3) the timing with which the required pickup orientation became visually specified.

## METHODS

*Design*

This study examined how different movement characteristics of a robotic effector influence temporal and spatial measures of human performance during robot-to-human object handovers. We present 3 experiments: Experiment 1 examined the effects of movement initiation and synchronization strategies, Experiment 2 compared different robotic effector velocity profiles, and Experiment 3 examined the effects of the timing with which task-relevant object information became available. A fourth experiment examining the effect of visual partner appearance (i.e., comparing human and robotic partner avatars) is presented in Supplementary Information. The purpose of these VR experiments was not to emulate a complete human-robot interaction context, but rather to isolate specific movement characteristics of a robotic effector and quantify their influence on human motor performance during object handovers. This experimental control allowed systematic manipulation of movement characteristics that would be difficult to achieve consistently and safely with physical robotic platforms.

*Participants*

Two separate samples of right-handed participants were recruited for the study. The first group (used in Experiment 1) consisted of 24 individuals with self-reported normal or corrected-to-normal vision and no history of neurological conditions. A group of 24 participants completed both Experiments 2 and 3; they self-reported to have normal or corrected-to-normal vision and no neurological conditions. Four participants in Experiment 1 (and 4 – see Supplementary Information) also participated in Experiments 2 and 3. Participants across both groups were primarily recruited through a university research participation scheme in exchange for course credit. Written informed consent was obtained prior to participation.

*Experimental setup*

A Virtual Environment as designed in Unity3D 2021.3.6f1 (Unity Technologies, San Francisco, CA, USA) and coded using Visual Studio 2019 (Microsoft Corporation, Redmond, WA, USA). The aim was to implement experimental conditions in a working and adjustable virtual robotic arm using an HTC VIVE Pro Eye Headset (HTC Corporation, Taoyuan City, Taiwan), three active Valve Index Base Station 2.0 units (Valve Corporation, Bellevue, WA, USA) operating through SteamVR (Valve Corporation, Bellevue, WA, USA). The environment comprised a simplified virtual room with flooring and walls to mimic physical surroundings. In the physical lab, participants were seated behind a wooden table, which had VIVE tracker 3.0 (HTC Corporation, Taoyuan City, Taiwan) attached to show a similar sized virtual table in the same position and orientation within the virtual environment. We expressed all positions in a table-centered coordinate system. The VIVE tracker 3.0 was utilized to track the position and rotation of the physical table in the lab and used to transform all positions and rotations to a coordinate system with the origin (0, 0, 0) at the near left corner of the table and axes oriented as follows: the positive x-axis was rightward, the positive z-axis was towards the robotic effector, and the positive y-axis was upwards (all from the participant's forward facing perspective).

A physical aluminum "grabber" object was utilized to pick up objects, accompanied by its virtual counterpart in the simulation (see Figure 1). The grabber featured an extended aluminum tip; its virtual version of which had a virtual 'magnet' embedded in its tip. This magnet was implemented using a 7mm distance threshold-dependent switch between the virtual cylindrical peg being attached to the robot arm and it being attached to the grabber tip. The pertinent distance between the center of the virtual magnet and the center top of the peg was continuously tracked for this purpose. Tracking of both physical and virtual objects was facilitated by a VIVE tracker (3.0) mounted on a horizontal bar attached to the Grabber, ensuring consistency between physical actions and their virtual representation. The horizontal bar with the tracker on top were not visible in the virtual environment to minimize visual occlusion of the virtual task (see Figure 1).

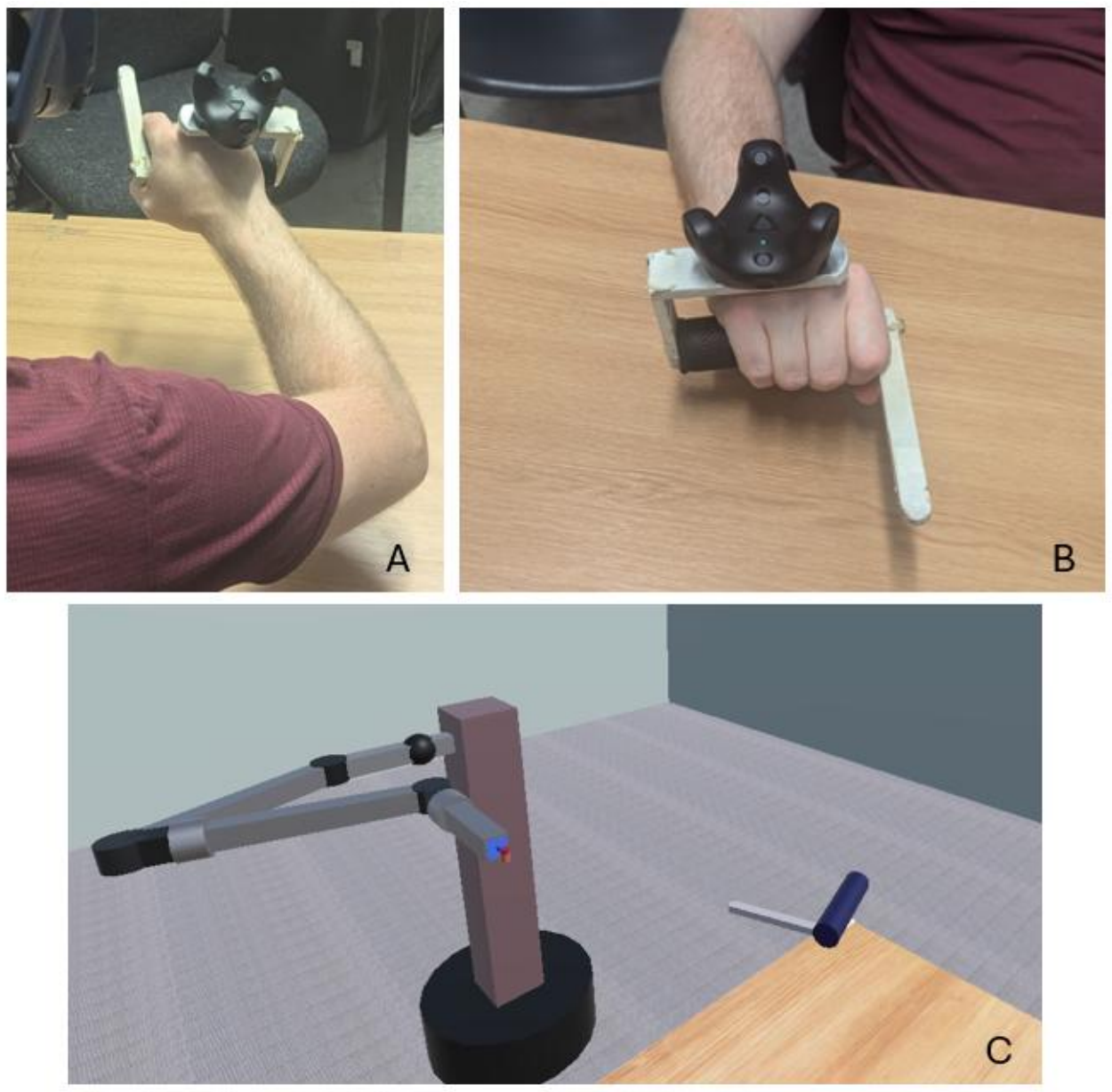

Figure 1. The physical grabber and its virtual counterpart. The grabber object in the real-world with both a view from behind the participant (A) and a top view facing the participant (B). Panel C shows the virtual view of the grabber and robotic effector.

The core of the simulation was a commercially available robotic arm Unity asset with inverse kinematics (Meuse Robotics[1]), which was extensively modified to implement the experimental conditions. The effector consisted of six joints and was visually and dimensionally modified to suit experimental requirements. A virtual base ensured proper alignment behind a virtual table for consistency in visual presentation (Figure 1C, 2). The effector's control was implemented using forward and inverse kinematics scripts provided with the asset. Forward kinematics calculated link and end-effector positions based on joint angles, while inverse kinematics employed the (Moore-Penrose) pseudoinverse of the Jacobian for iterative joint updates. Joint angles were constrained to a range of -160° to 160° - no updates were made if any new joint angle exceeded this range. An additional joint was added – akin the 'shoulder' of the robotic arm – rotations of which were only used to change the orientation of the entire robot arm (i.e., in the Supplementary experiment – see Supplementary Information).

[1] Original asset available from the Unity Asset Store: *Robotic Arm with Inverse Kinematics* (Meuse Robotics), accessed 19 February 2025. https://assetstore.unity.com/packages/templates/packs/robotic-arm-with-inverse-kinematics-161155

A default smooth movement trajectory was implemented for the robot arm using a minimum jerk criterion (which captures key features of human movements; [27]), although in some experiments different trajectories/mechanisms were used (as will be described in the respective experiments). The minimum jerk trajectory was generated using a function that implemented polynomial equations [59].

$$C3 = (-20 \cdot X_0 + 20 \cdot X_E)/(2 \cdot MT^3) \quad (1)$$

$$C4 = (15 \cdot X_0 - 15 \cdot X_E)/MT^4 \quad (2)$$

$$C5 = (-12 \cdot X_0 + 12 \cdot X_E)/(2 \cdot MT^5) \quad (3)$$

Here, *C3, C4,* and *C5* are coefficients that are calculated based on the initial ($X_0$) and final ($X_E$) positions, and the duration of the movement (*MT*).

$$position(t) = X_0 + t^3 \cdot C3 + t^4 \cdot C4 + t^5 \cdot C5 \quad (4)$$

$$velocity(t) = 3 \cdot t^2 \cdot C3 + 4 \cdot t^3 \cdot C4 + 5 \cdot t^4 \cdot C5 \quad (5)$$

$$acceleration(t) = 6 \cdot t \cdot C3 + 12 \cdot t^2 \cdot C4 + 20 \cdot t^3 \cdot C5 \quad (6)$$

Equations (4), (5), and (6) express the position, velocity and acceleration based on the previously calculated coefficients and the time variable (*t*, relative to movement initiation). This allowed calculation of the kinematics of the peg attached to the robot end-effector at any time following movement initiation.

All coordinates for data analyses were expressed in a replicable table-centered system (with the origin at the near left table corner and the positive x-axis pointing rightward, the positive y-axis pointing upward, and the positive z-axis pointing forward toward the robot). A VIVE tracker 3.0 recorded the table's position and orientation in the physical room. Transformations between world- and table-centered coordinates were applied using Unity3D’s quaternion operations.

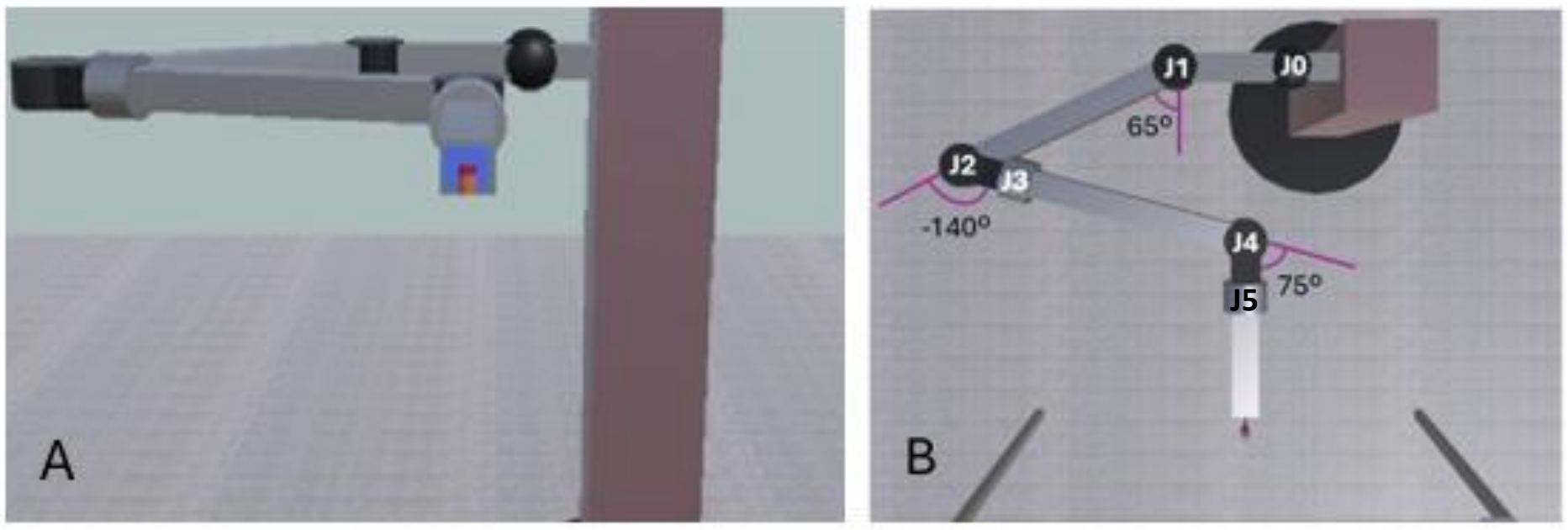

Figure 2. Virtual robot arm. Participant view of the robot arm used for this study set to a 90° ‘elbow’ elevation (A). An aerial view including initial joint configurations for the six joints of the robot arm (J0 = 0°; J1 = 65°, J2 = -140°; J3 = 0°; J4 = 75°; J5 = 0°; B). Note: J0, while a sphere, only allows rotation around the depth axis (z-axis in our ‘table centered’ framework).

Before each trial, our custom code set the position, orientation, and visibility of all objects as appropriate. Trials then started when participants touched top of a cylinder, which also was the receptacle for subsequent peg drop off, with the tip of the virtual grabber. The virtual grabber tip had to be within 7mm of the top centre of the virtual receptacle for at least 200ms consecutively: when the criteria had been met a beep sound played to signal trial onset. The participant then moved the grabber to a virtual peg attached to the robot arm's end-effector. The task required participants to pick up the peg using virtual magnet in the grabber's tip (Figure 3A). To achieve seamless visual transitions during peg pickup, two identical virtual pegs were used: one attached to the robot and one to the grabber. The robot-held peg was visible at trial onset and upon satisfying the pickup criteria, it was deactivated while the Grabber-held peg became active, simulating a magnetic snap. Once the robot arm retreated, participants moved the grabber plus peg (if picked up) to a designated receptacle at the front of the table (Figure 3). For trials with successful pick-ups, peg drop-off occurred when the peg centre came within 7mm of the top centre of the receptacle. Trials concluded once the virtual magnet was within came within 7mm of the top centre of the receptacle, and the robot arm had reached its starting position.

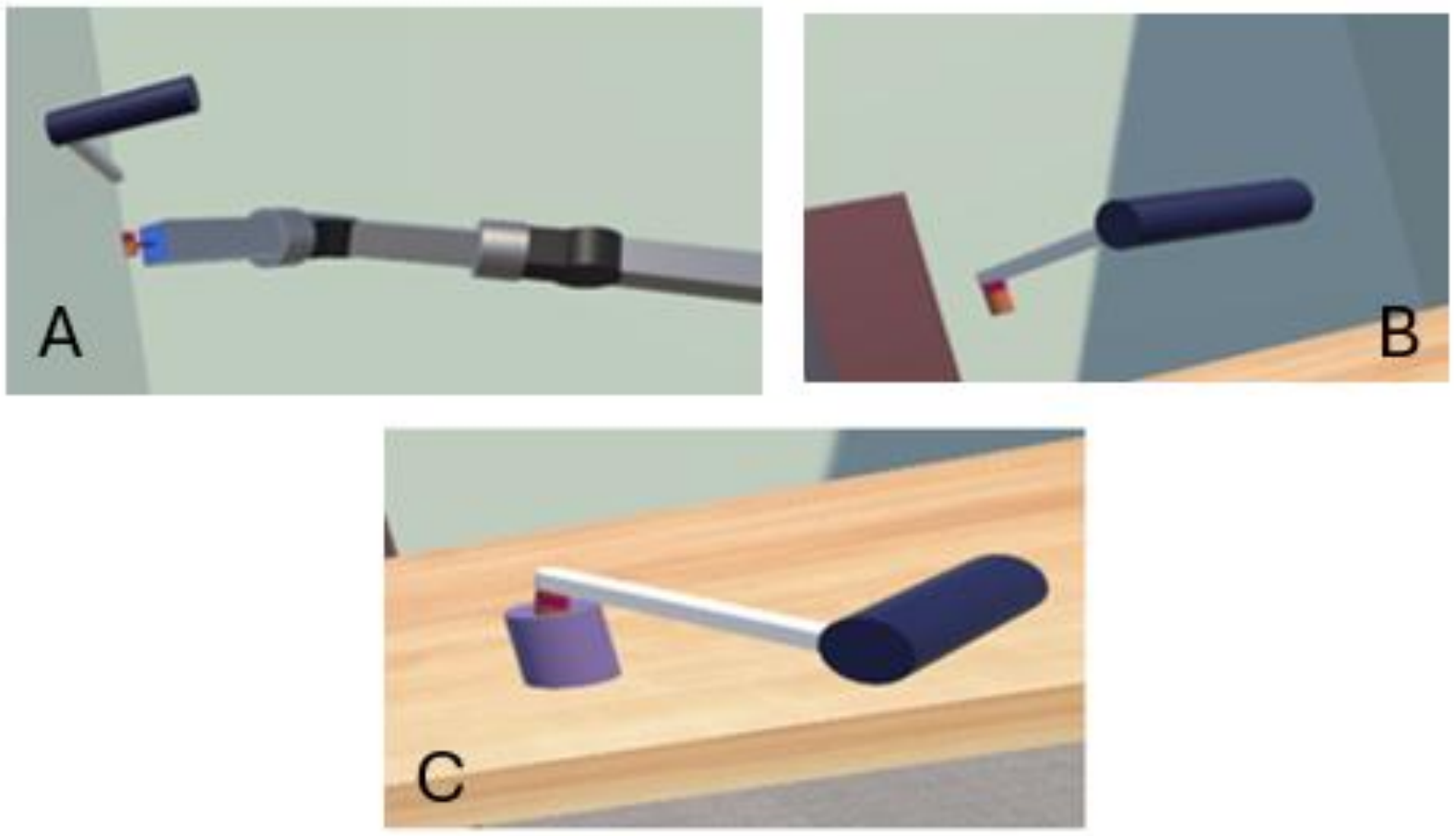

Figure 3. General task sequence. Participants held and moved the grabber, the tip of which acted as a virtual magnet that attracted the top of an orange and pink-tipped peg, snapping it in place once close enough (A). Once the peg was attached, the participant moved the grabber and peg back to the purple peg receptacle (B), while the robot arm returned to its starting position. Participants placed any picked up pegs into the peg receptacle at the end of each trial (see text for details), which ended the current trial and initiated the next (also acting as the trial starting position; C).

*Experiment 1*

Participants completed four conditions with different interaction algorithms that were utilized by the robotic effector to move the peg 0.35m forward. This included a *robot-initiation* condition, in which the robot started moving its end-effector along the predefined minimal jerk trajectory for 1000ms directly following trial onset, with participants responding to this. Secondly, in the *participant-initiation* condition the robot only starting to move once the participants had moved the grabber tip 10mm from the initial position. In a third *temporal-alignment* condition, the robot controlled the peg trajectory online to synchronize arrival of the peg at the desired final position with arrival of the grabber tip at a predefined interaction depth ($Z_{int}$) midway between the initial z-coordinates of the receptacle and the peg (which were 70cm apart). This was achieved by dynamically adjusting the forward peg position based on the relative position of the Grabber tip:

$$\boldsymbol{Pos_{peg}}(t) = \boldsymbol{Pos_{peg}}(0) + (\boldsymbol{Pos_{int}} - \boldsymbol{Pos_{peg}}(0)) \cdot \frac{\left(Z_{tip}(t) - Z_{receptacle}(0)\right)}{\left(Z_{int} - Z_{receptacle}(0)\right)} \quad (7)$$

Here, $\boldsymbol{Pos_{peg}}$ is the (three-dimensional [3D]) position of the peg at the current time (*t*). Equation 7 ensures the robot (i.e., peg) at any frame is positioned at the same relative distance (in forward, Z, direction) from its initial position to the interaction depth ($Z_{int}$) as the grabber tip. $\boldsymbol{Pos_{int}}$ is the predefined 3D end position of the peg. This control meant that the robot did not move until the participant initiated their movement. Updating based on this equation was stopped in the last 3% of the Z distance to be covered, with the robot motion stopping to mitigate any detrimental effects of participant overshoot. For this and the prior two algorithms the peg was moved to one of four end points (positioned on a two-by-two grid of upper left, upper right, lower left, and lower right: all relative to the robot-held pegs initial position).

A fourth *spatiotemporal-alignment* condition also included real-time control of the peg position by the robot, but did not constrain this to fall along a straight path. Rather, the peg position was adjusted to mirror the tip position (with an additional correction for the difference between the initial positions). Effectively, this meant the robot moved the peg to wherever the participant chose to move the tip, using kinematics closely matching that of the participant.

$$\boldsymbol{Pos_{peg}}(t) = \boldsymbol{Pos_{peg}}(0) + \boldsymbol{D}\left(\boldsymbol{Pos_{tip}}(t) - \boldsymbol{Pos_{receptacle}}(0)\right) - \boldsymbol{Offset}(t) \quad (8)$$

Equation (8) describes the updating of the coordinates of the peg position at the current time (*t*). ***D*** is the movement direction of [1 1 –1], which means the robot kinematics match the participant's motion direction in the x- and y-axis and reverse this direction in the z-axis: as the grabber moves in positive direction (away from the participant), the peg is moved in a negative direction (towards the participant). The initial peg receptacle position is subtracted from the grabber tip position, correcting for any imperfect initial tip positions in the x- and y-direction (as the distance criterion between the grabber tip and peg receptacle allows for slight deviations upon trial onset). The relative movement of the grabber tip position is then corrected for required offsets:

$$\boldsymbol{Offset}(t) = \left(\boldsymbol{Pos_{peg}}(0) - \boldsymbol{Pos_{receptacle}}(0)\right) \cdot \frac{\left(Z_{tip}(t) - Z_{receptacle}(0)\right)}{\left(Z_{int} - Z_{receptacle}(0)\right)} + \begin{bmatrix} 0 \\ 0.04 \\ 0 \end{bmatrix} \quad (9)$$

Over the course of the forward motion, the offset (9) considers the difference between the initial position of the peg and the initial receptacle position (where the grabber tip starts), to correct for positional offsets at trial onset, including a(n approximate) 22cm y-axis offset between the initiation point and the peg. It multiplies this difference by the relative displacement in the z-direction to adjust for the relative movement in this direction. An additional 4cm offset was applied at the beginning of the trial in the y-direction to ensure the robot was always moving to a position 4cm below the Grabber tip. The updating was stopped in the last 3% of the distance to be covered, after which the robot motion stopped to allow the participant to pick up the peg. For both *temporal-alignment* and *spatiotemporal-alignment*, once the participant had moved 20mm away from the robot tip after attempting a pick up, the robot began the return movement back to its initial position by matching the return of the peg using the same equations described above. In both the *participant-initiation* and *robot-initiation*, the robot simply reversed its forward trajectory, but at 1.5x its original speed, to return to its initial position. The robot moved the peg to one of four final X-Y positions on a two-by-two grid of upper left (-0.15m [x], 0.15m [y]), upper right (0.15m [x], 0.15m [y]), lower left (-0.15m [x], -0.15m [y]), and lower right (0.15m [x], 0.15m [y]).

*Experiment 2*

In Experiment 2 we varied the forward velocity profile of the robot arm. These profiles were categorized into four types (see Figure 4): *minimum-jerk, constant-velocity, constant-acceleration,* and *biphasic*. Importantly, this included the addition of trial-to-trial random variations in the final depth of the peg to prevent predictability (i.e., in each trial we added a random value between -0.02m and 0.02m to the fixed final depth).

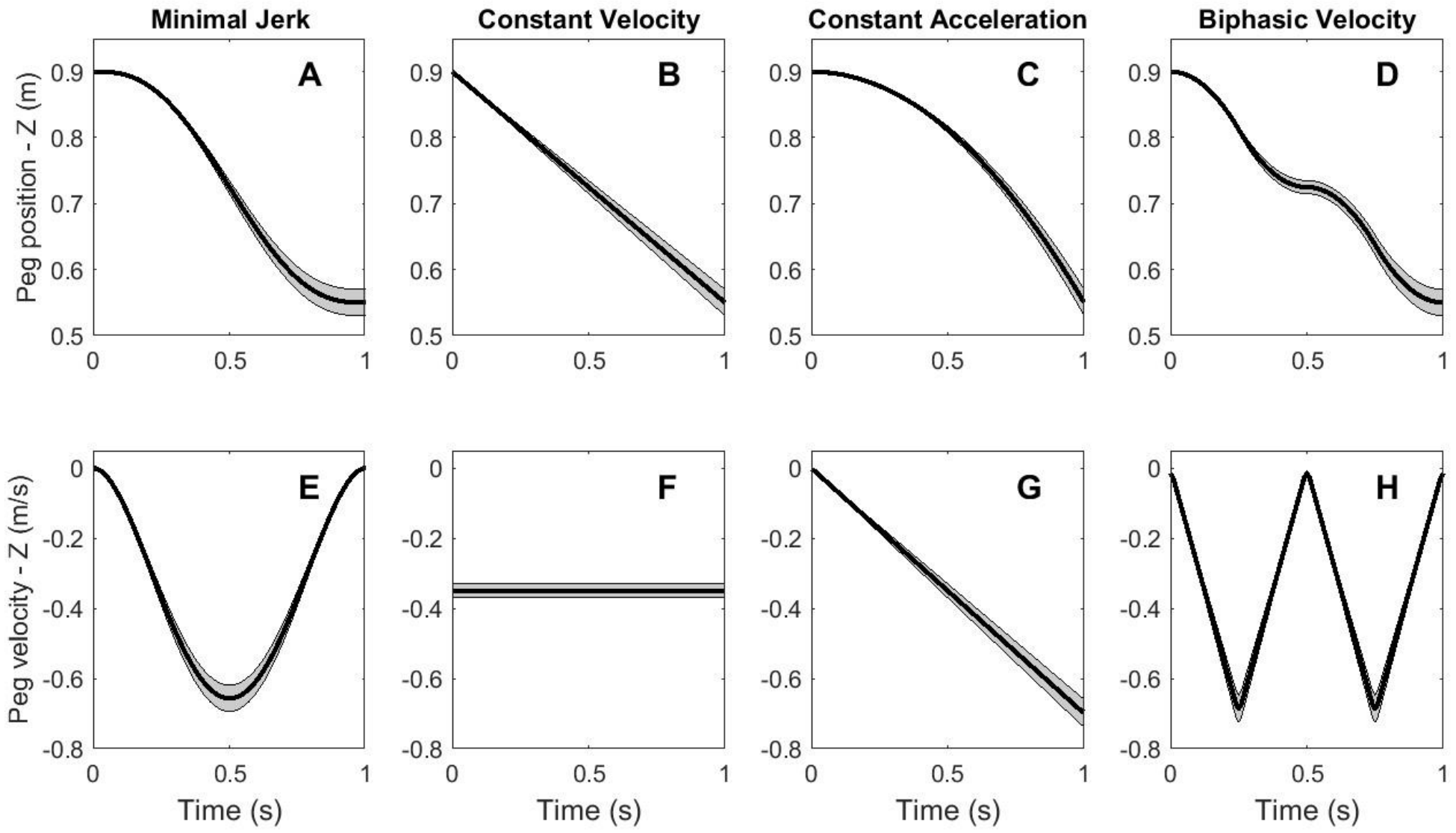

Figure 4. Manipulation of the robot kinematics: *minimum-jerk* position (A) and velocity profiles (E), *constant-velocity* position (B) and velocity profiles (F), *constant-acceleration* position (C) and velocity profiles (G), and *biphasic* position (D) and velocity profiles (H). Note that all panels show a grey shaded region, which represents the range of kinematics possible due to variations in the final depth (-2cm to 2cm) the robot moved the peg to.

The *minimum-jerk* trajectory utilized in the first condition follows the default trajectory implemented Experiment 1 (Equations 1-6), plus the aforementioned random variation in final depth. In the *constant-velocity* condition the robot moves the peg from an initial position ($\boldsymbol{Pos_0}$, a 3D position vector) to a final position ($\boldsymbol{Pos_E}$, a 3D position vector) with a constant speed throughout the total movement time (*MT*). The 3D peg position at time (*t*) ($\boldsymbol{Pos}(t)$) is given by:

$$\boldsymbol{Pos}(t) = \boldsymbol{Pos}_0 + \frac{(\boldsymbol{Pos_E} - \boldsymbol{Pos_0})}{MT} \cdot t \quad (10)$$

For *constant-acceleration* the peg trajectory consists of uniformly accelerated motion from the initial position ($\boldsymbol{Pos_0}$) to the final position ($\boldsymbol{Pos_E}$), starting with a speed of 0:

$$\boldsymbol{Pos}(t) = \frac{1}{2} \cdot \frac{2 \cdot (\boldsymbol{Pos_E} - \boldsymbol{Pos_0})}{MT^2} \cdot t^2 + \boldsymbol{Pos}_0 \quad (11)$$

The biphasic velocity profile was calculated in four consecutive stages of constant acceleration (12), deceleration (13), acceleration (14) and deceleration (15), all of equal duration:

For $0 \leq t < 0.25$: $\boldsymbol{Pos}(t) = \boldsymbol{Pos}_0 + \frac{4 \cdot (\boldsymbol{Pos_E} - \boldsymbol{Pos_0})}{MT^2} \cdot t^2$ (12)

For $0.25 \leq t < 0.5$: $\boldsymbol{Pos}(t) = \boldsymbol{Pos_0} + \frac{1}{4} \cdot (\boldsymbol{Pos_E} - \boldsymbol{Pos_0}) + \left(t - \frac{MT}{4}\right) \cdot \left(\frac{2 \cdot (\boldsymbol{Pos_E} - \boldsymbol{Pos_0})}{MT}\right) - \frac{4 \cdot (\boldsymbol{Pos_E} - \boldsymbol{Pos_0})}{MT^2} \cdot \left(t - \frac{MT}{4}\right)^2$ (13)

For $0.5 \leq t < 0.75$: $\boldsymbol{Pos}(t) = \boldsymbol{Pos_0} + \frac{1}{2} \cdot (\boldsymbol{Pos_E} - \boldsymbol{Pos_0}) + \frac{4 \cdot (\boldsymbol{Pos_E} - \boldsymbol{Pos_0})}{MT^2} \cdot (t - \frac{MT}{2})^2$ (14)

For $0.75 \leq t < 1.0$: $\boldsymbol{Pos}(t) = \boldsymbol{Pos_0} + \frac{3}{4} \cdot (\boldsymbol{Pos_E} - \boldsymbol{Pos_0}) + \left(t - \frac{3 \cdot MT}{4}\right) \cdot \left(\frac{2 \cdot (\boldsymbol{Pos_E} - \boldsymbol{Pos_0})}{MT}\right) - \frac{4 \cdot (\boldsymbol{Pos_E} - \boldsymbol{Pos_0})}{MT^2} \cdot \left(t - \frac{3 \cdot MT}{4}\right)^2$ (15)

To better match the physics of magnets, the grabber-peg interaction was extended for Experiments 2 and 3 with an additional orientation criterion: peg pickup was only possible when the two surfaces had a similar enough orientation (i.e., peg pickup was only possible if the angular difference did not exceed 10° for both pertinent cardinal axes of the peg). Additionally, to promote more synchronized partner movements (i.e., to avoid the participant just waiting to move until the robot reached its end posture), the robot only kept the peg at its final location for 550ms before initiating its return movement. The robot moved the peg to one of four final X-Y positions on a two-by-two grid of upper left (-0.15m [x], 0.1m [y]), upper right (0.15m [x], 0.1m [y]), lower left (-0.15m [x], -0.1m [y]), and lower right (0.15m [x], 0.1m [y]) – this minor difference with Experiment 1 was motivated to avoid fatigue associated with the higher final positions.

*Experiment 3*

In Experiment 3, we varied the timing of the peg rotation around the depth (z) axis during the robot's translational movement, using four variations. In the *early-rotation* and *late-rotation* conditions, the rotation of the peg occurred in the first or final half of the robot's translational movement (early: rotation start = 0ms*;* rotation end = 600ms; late: rotation start = 600ms*;* rotation end = 1200ms). In the *synced-rotation* condition the robot rotated the peg throughout its translational movement (rotation start = 0ms; rotation end = 1200ms). The *pretrial-rotation* condition, was included as a control condition. In this condition, the robot rotated the peg prior to its translation movement, ensuring their forward translation started with the relevant end orientation (rotation start

= -400ms*;* rotation end = 0ms). Peg rotation and translation were implemented as positional and angular minimum jerk trajectories, respectively. The peg had two possible initial orientations (70° or 110°) and two possible end orientations (-20° or 20°), relative to vertical; from the participant's perspective the peg always rotated in the clockwise direction. This experiment used the same 2x2 grid of final X-Y peg positions and small random variations in the Z dimension as Experiment 2 (see above).

***Procedures***

Participants arrived in the lab and were taken through the ethics aspects of their participation, explained the general purpose of the experiment(s) and the use of virtual reality equipment, and provided the opportunity to ask any questions. Once satisfied, they signed an informed consent form. Each experiment consisted of a structured sequence of handover trials, typically divided into blocks separated by short breaks, as outlined below for each experiment. Practice trials were included at the beginning of each block or experiment to ensure participants were familiar with the VR environment and task. The duration of each experiment was approximately 45 minutes.

In Experiment 1, the three primary conditions were presented in blocks, with the order counterbalanced across the 24 participants. The exploratory *spatiotemporal-alignment* condition was always completed as the final, fourth block. In each block, we included 24 repetitions of each of the four final positions. The total number of trials across the four blocks thus was 384 ([24 x 4] x 4 = 384), although each block also started with 10 familiarization trials to ensure the participant was comfortable with the task and environment (15 in the first block and 10 in subsequent blocks).

In Experiment 2, 24 repetitions of the peg's 4 velocity profile and 4 final positions (384 trials in total) were presented in fully randomized order across four blocks separated by short breaks. Each block started with five practice trials for (re)familiarization with the environment (with randomly selected conditions); 10 extra practice trials were included in the first block.

In Experiment 3, the four timing conditions were presented in separate blocks, the order of which was randomized for all participants. For each timing condition, there were 16 conditions (2 initial peg orientations x 2 final peg orientations x 4 final positions, see above). Six repetitions of these conditions (resulting in 96 repetitions for each timing variation) were presented in fully randomized order. Again, there were (re)familiarization trials at the start of each block, as in Experiment 3 (see above).

This study received ethical approval from Queen's University Belfast Engineering and Physical Science Faculty Research Ethics Committee (reference number EPS 23_10).

***Data analyses***

For each trial, we saved the 3D position and orientation (quaternions) data of pertinent objects in the environment as well as pertinent state variables for each frame. All positional data was expressed in a coordinate system attached to the table (origin at the near left corner). Preprocessing was completed in MATLAB (The MathWorks, Inc., Natick, MA, USA); analyses focused on the positional data. We first resampled the data to exactly 90Hz using cubic spline interpolation based on the exact time of the frame presentation. The forward (z) position signal was visually inspected for each trial to identify signal deviations (i.e., abrupt drops or spikes in the signal); these trials were excluded from the analyses as deviations were judged to most likely reflect tracking errors and not natural movement variations. The retained data were filtered using a recursive low-pass fourth-order Butterworth filter (10Hz cut-off). Grabber tip velocity along the z-axis was calculated using the gradient function in MATLAB.

Visual signal deviations resulted in the exclusion of 18 trials from Experiment 1 and 1 trial for Experiment 2 (no visual deviations were observed in Experiment 3). Additional exclusions were made in Experiments 2 and 3 based on evidence of trial disengagement, which was defined as trials in which the grabber tip depth did not get within 50mm of the final peg depth before the robot started moving back. This led to the exclusion of 8 trials in Experiment 2 and 3 trials in Experiment 3. In addition, participants who missed the peg-pickup for more than 50% of the trials (in Experiment 2 or 3) were excluded from the data analysis; this resulted in the exclusion of five participants for Experiment 2 and 3.

The start of the pickup movement was defined based on several criteria: (1) the z-position of the grabber tip had to be less than 35cm from the front edge of the table; (2) the grabber tip must have covered less than 20% of the to-be-moved forward (z) distance; and (3) the grabber tip's z-velocity exceeded a threshold of 3% of its maximum. The first sample where all these criteria were met was taken as the pickup initiation. If these criteria were not met, pickup initiation was determined as the first sample where the velocity exceeds the threshold set - this was to handle instances where the participant was in motion before the onset of the trial. The end of the pickup movement was defined as the moment the grabber tip's z-position came within 1cm of either the final z-position of the robot-held peg or its own maximal forward (z-)position, whichever came first. Peg pickup was defined to occur when the grabber tip came within 7mm of the center top of the peg (plus an orientation criterion in Experiments 2 and 3). Peg Pickup Success coded whether peg pickup occurred during a trial or not.

The drop-off movement was defined slightly differently, mainly as a result of the occasional presence of multiple velocity peaks in the data associated with the slightly different nature of the movement. First, we defined grabber tip z-velocity minima lower than -0.2m/s (i.e., during the backward movement) from the onset of the robot return movement until the end of the trial. If no such minima were found in this window, drop-off initiation was defined as the sample following the last time before the last z-velocity minimum lower than -0.2 m/s the z-velocity dropped below 5% of the minimal z-velocity. If minima were found, we determined whether there were zero-crossings in the grabber tip z-velocity profile between the robot return sample and the first z-velocity minimum smaller than -0.2m/s. If zero-crossings were found, we offset the grabber tip z-velocity by the average grabber tip z-velocity in the 100ms prior to the robot return onset, but only if this average was negative. Drop-off initiation was defined as the first sample (from 100ms prior to the robot return onset until the first velocity z-minimum smaller than -0.2ms) the grabber tip z-velocity (offset or not) was lower than 5% of the minimal z-velocity. The end of the drop-off movement was defined as the first time after the start of the drop-off movement the grabber tip z-position was within 35cm from the front edge of the table, and the grabber tip z-velocity was higher (i.e., less negative) than 5% of the minimal grabber tip z-velocity during the trial. In cases where the z-velocity never became higher than this criterion (i.e., when the drop-off occurred with the grabber tip in motion), the last sample in the trial was taken as the end of the drop-off movement.

We calculated variables associated with different phases of the movement: the pickup phase, the manipulation or interaction phase, and the dropoff phase. Dependent variables considered across experiments were pickup initiation, movement durations (pickup and drop-off), overall movement duration, total trial duration, manipulation duration (between the end of the forward movement and the successful pickup of the peg), interaction duration (between the end of the forward movement and the start of the return movement), the endpoint of the pickup movement relative to the final peg position (lateral [x-axis], vertical [y-axis], and depth [z-axis]), drop-off reaction time, path lengths (for the pickup, manipulation, interaction [only in Experiments 2 and 3], and drop-off phases), and peg-pickup success (also only in Experiments 2 and 3). Full details of the exact calculations are presented below in Table 1.

For each experiment, the Results section focuses on the effects considered most relevant for interpreting human motor performance during the handover task (see Supplementary Information, Tables S1–S5 for the complete set of results). This selective reporting in the main text was adopted to maintain clarity and is appropriate given the exploratory nature of the study. Specifically, effects only presented in the Supplementary tables the for all experiments are trial durations and total movement durations, alongside interaction duration and interaction path lengths for Experiments 2 and 3. The relative pickup endpoint effects for the spatiotemporal-alignment algorithm of Experiment 1 are also not presented in the results (these measures were determined to be more reflective of the algorithm's configuration to mirror participants' positional adjustments). Importantly, our statistical approach (described in detail below) considered all tests conducted. The selective discussion of results in the main text was therefore motivated by clarity of presentation rather than differential statistical treatment of the variables.

Table 1: Dependent Variables defined and examined in the Data Analysis

| Dependent Variable | Definition |
|---|---|
| Time of Initiation | The time at which the pickup movement starts. |
| Trial duration | The time relative to trial onset at which the drop-off phase ends. |
| Movement duration | The difference between the end of the drop-off phase and the start of the pickup phase. |
| Movement durations for pickup and drop-off phases | The time differences between the start and end of each phase. |
| Movement duration of manipulation phase | The time between the end of the pickup phase and the moment the peg is picked up. For Experiments 2 and 3 this only encompasses trials with a successful pickup. |
| Movement duration of interaction phase | The time between the end of the pickup phase and the start of the drop-off phase. This was calculated for all trials (i.e., even when the pickup was unsuccessful in Experiments 2 and 3). |
| Reaction time for the drop-off movement | The time between the moment the robotic effector starts moving back (which is when the peg was picked up on successful trials) and the start of the drop-off phase. |
| Symmetry of velocity profiles for the pickup and drop-off phases | The duration of the acceleration phase (between movement onset and peak velocity) relative to the duration of the phase. Values smaller than 0.5 imply the acceleration phase (between the peak and offset) is shorter than the deceleration phase. |
| Path length | The superfluous displacement, calculated separately for the pickup, manipulation, interaction, and drop-off phases, respectively. The path length itself is the cumulative sum of between-sample grabber tip 3D displacements from onset to offset of the movement phase. The 3D distance between the initial and final sample of each phase was subtracted to capture the superfluous displacement. |
| Relative pickup endpoint in the x, y and z-axis | The 3D position of the grabber tip at the end of the pickup movement relative to the top of the peg at this sample. |
| Peg-pickup success | The percentage of successful pickups across valid trials (per condition) for Experiments 2 and 3 (when the robotic effect could move back before any pickup). |

*Statistical analyses*

Within-subjects General Linear Models (GLM) were used in SPSS to compare differences in the variables between the pertinent conditions. When the sphericity assumption was violated (Mauchly's test significant [alpha = 0.05]), Greenhouse-Geisser [30] corrected degrees of freedom were used, although the Huynh-Feldt [34] correction was used except when epsilon exceeded 0.75 [26] We used partial eta-squared ($\eta_p^2$) as a measure of effect size for parametric data, to afford comparing effect sizes between experiments and studies [36, 48]. Non-parametric tests (Friedman tests) were used when normality could not be assumed for more than one of the levels (as determined by Shapiro-Wilk tests and using a cutoff of $\alpha < 0.05/N$ (where N is the number of levels, which was 4 for all experiments [8]). Kendall's *W* was used as a measure of effect size for non-parametric data. Variables with large effect sizes ($\eta_p^2 \geq 0.25$ [32] or $W \geq 0.5$ [12] were further explored during post-hoc analyses. Critically, none of the decisions to conduct follow-up tests depended on statistical significance levels, reflecting the exploratory nature of the study.

All follow-up tests for large effects were between-condition comparisons and involved paired-samples *t*-tests for data matching parametric assumptions or Wilcoxon signed-rank tests otherwise. Šidák [58] corrections were applied to the *p*-values for the total number of post hoc tests conducted ($p_s$=1-(1-$p$)$^k$), where $p_s$ is the Šidák-corrected *p*-value, *p* is the raw uncorrected *p*-value, and *k* is the number of large effects × the number of post hoc comparisons. In Experiment 1, the family-wise error correction was applied with *k* = 66, based on 11 large effects followed by 6 post hoc comparisons (11 × 6 = 66). In Experiment 2, *k* = 90 was applied based on 15 large effects each followed by 6 post hoc comparisons (15 × 6 = 90). In Experiment 3, *k* = 24 was applied, calculated as 4 large effects each with 6 post hoc comparisons (4 × 6 = 24).

Effects were selected for further consideration based on an effect size criterion. Note that all anonymized data and analysis code are available online (https://osf.io/5qvtx/), along with the raw, uncorrected *p*-values for completeness (n.b., no correction for multiple comparisons was applied because the *p*-values of the considered effects did not inform any statistical decisions).

## RESULTS

### *Experiment 1: Spatiotemporal Control of Object Handovers*

The goal of this experiment was to assess how movement initiation and temporal coordination influenced human performance during the robot-to-human handover task. The initiation conditions and temporal synchronization algorithms produced multiple large effects on movement behavior and task performance. Some of these differences are apparent in the raw data, which also clearly reflect features of the experimental manipulations (see Figure 5). For instance, robot-initiated motion (panel A) elicited earlier movement onsets and faster, more direct pickup trajectories. Panels C and D also highlight the temporal- and spatiotemporal-alignment implemented by the algorithms, which for both yielded a Z-trajectory of the robot-held peg that mirrored the Z-trajectory of the participant-held grabber's tip .

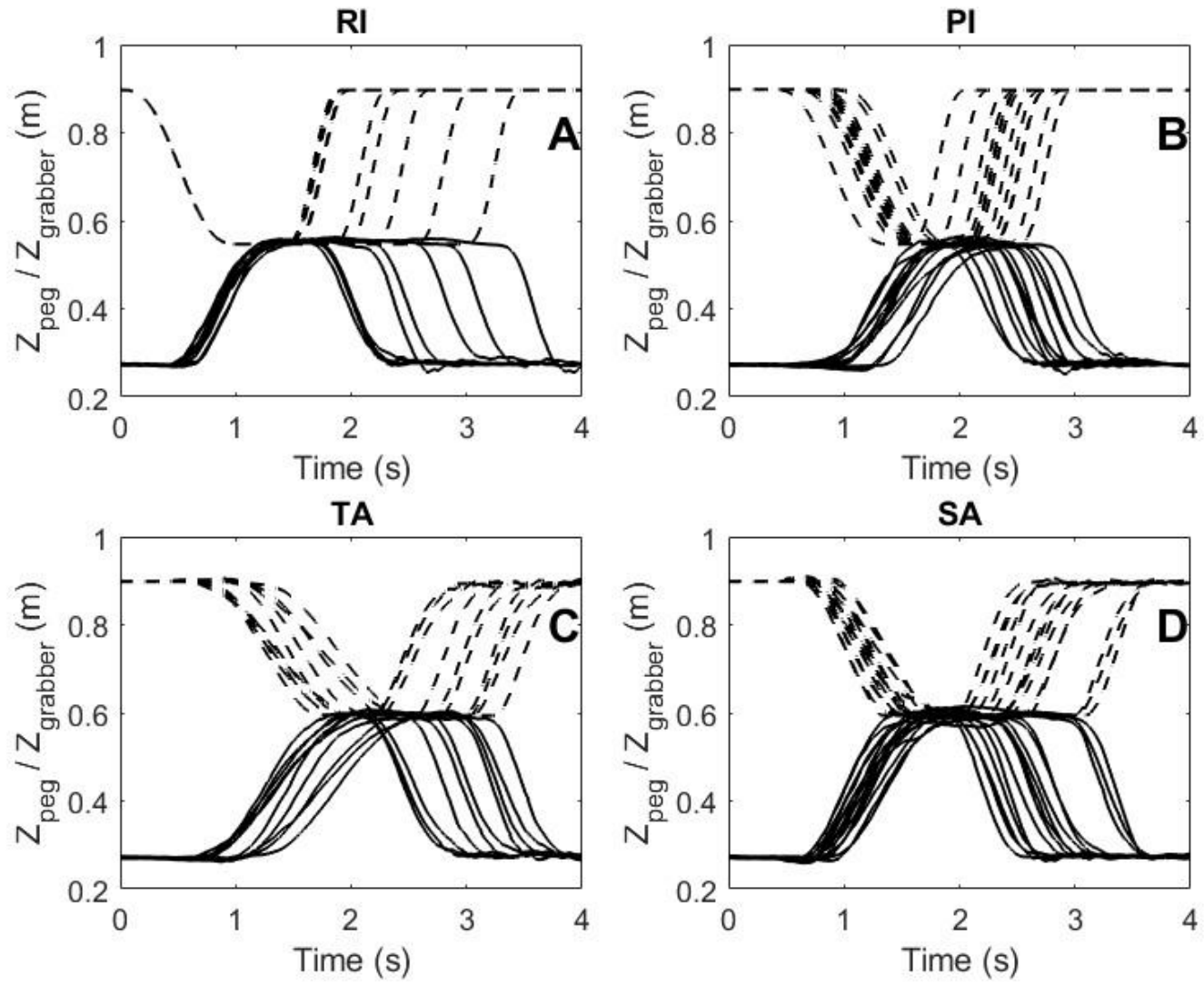

Figure 5. Raw Z-trajectories for the trials of one participant, for the peg held by the robotic effector (dashed lines) and the tip of the participant-held grabber (solid lines) illustrating differences between *robot-initiation* (RI, panel A), *participant-initiation* (PI, panel B)*, temporal-alignment* (TA, panel C), and *spatiotemporal alignment* (SA, panel D).

The relevant effects primarily reflected differences between the *robot-initiation* condition and the other interaction algorithms, as well as patterns associated with the *spatiotemporal-alignment* condition (Figure 6). Specifically, post hoc analyses revealed that pickup initiation occurred earlier for *robot-initiation* compared to *participant-initiation*. Moreover, *robot-initiation* facilitated shorter pickup movements and straighter pickup paths compared to both *participant-initiation* and *temporal-alignment* conditions. Additionally, pickup movements ended closer to the object in depth (i.e., z-axis), and pickup velocity profiles were more symmetrical under *robot-initiation* relative to *temporal-alignment*.

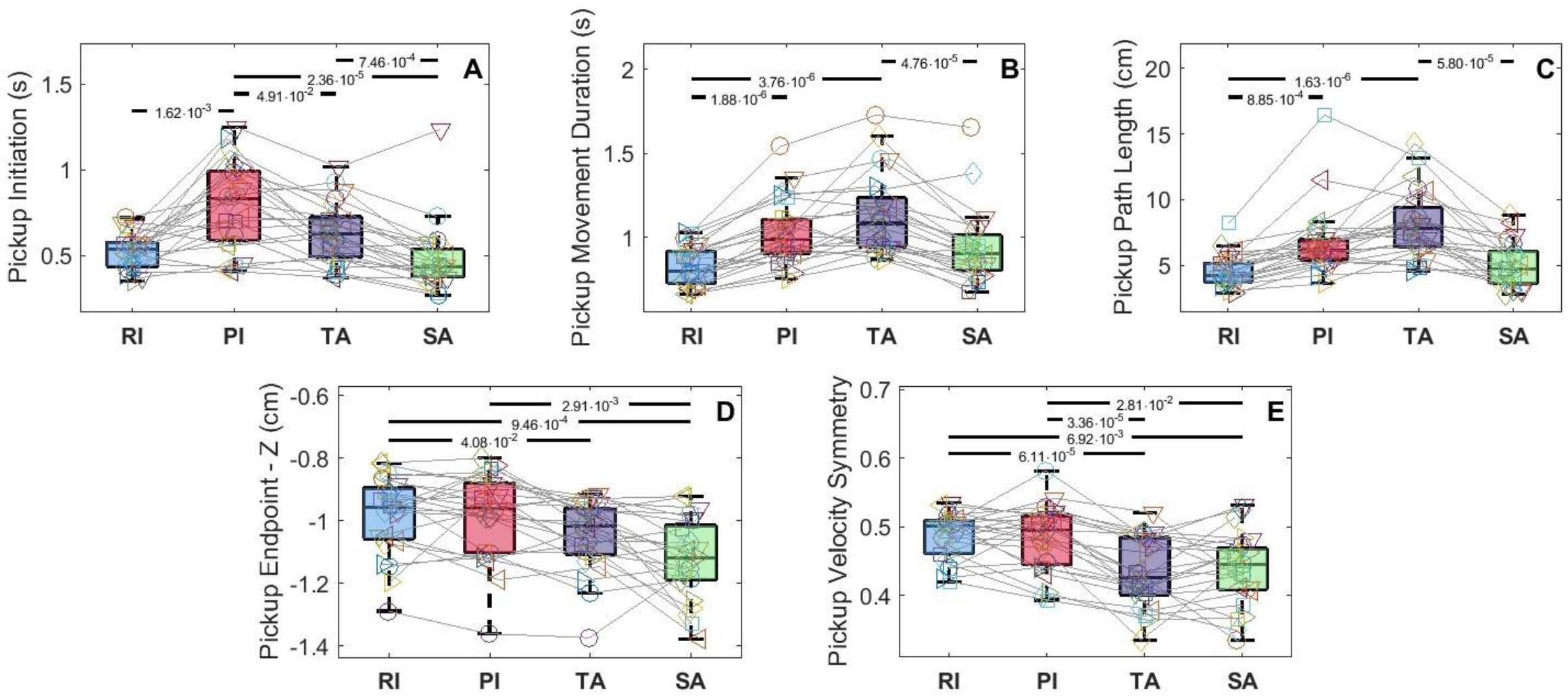

Figure 6. Large effects of interaction algorithm in Experiment 1. Differences in pickup initiation (A, $\eta_p^2$ = 0.53), pickup movement duration (B, $\eta_p^2$ = 0.58), pickup path length (C, $\eta_p^2$ = 0.53), pickup endpoint-Z (D, $\eta_p^2$ = 0.42), and velocity symmetry of the pickup movement (E, $\eta_p^2$ = 0.45) between *robot-initiation* (RI), *participant-initiation* (PI)*, temporal-alignment* (TA), and *spatiotemporal alignment* (SA). Significant follow-up differences are indicated by lines above the boxes, with the Šidák-corrected *p*-value embedded in the line. Note that a complete list of large effects and their statistics can be found in Supplementary Table S1.

Movements in the *spatiotemporal-alignment* condition differed from the *participant-initiation* and *temporal-alignment* conditions in several respects. Even though the robot in these conditions always began its main movement only after participant initiation, movements for the *spatiotemporal-alignment* condition were initiated earlier (also resulting in a shorter duration compared to the *temporal-alignment* condition). Further, pickup initiations was later and velocity profiles were more symmetrical with *participant-initiation* compared to *temporal-alignment*. Velocity profiles in the *spatiotemporal-alignment* algorithm pickups were less symmetrical compared to both *robot-initiatio*n and *participant-initiation* conditions, with longer pickup path lengths for *temporal-alignment*. Unexpectedly, *spatiotemporal-alignment* also influenced the drop-off phase (Figure 7), leading to longer drop-off path lengths compared to *robot-initiation* and *participant-initiation*, longer drop-off durations relative to *robot-initiation*, and more symmetrical velocity profiles for drop-off movements than all other conditions. Despite return movements being entirely participant-controlled across conditions, we thus also found that drop-off movements were quicker for *robot-initiation* compared to *temporal-alignment*.

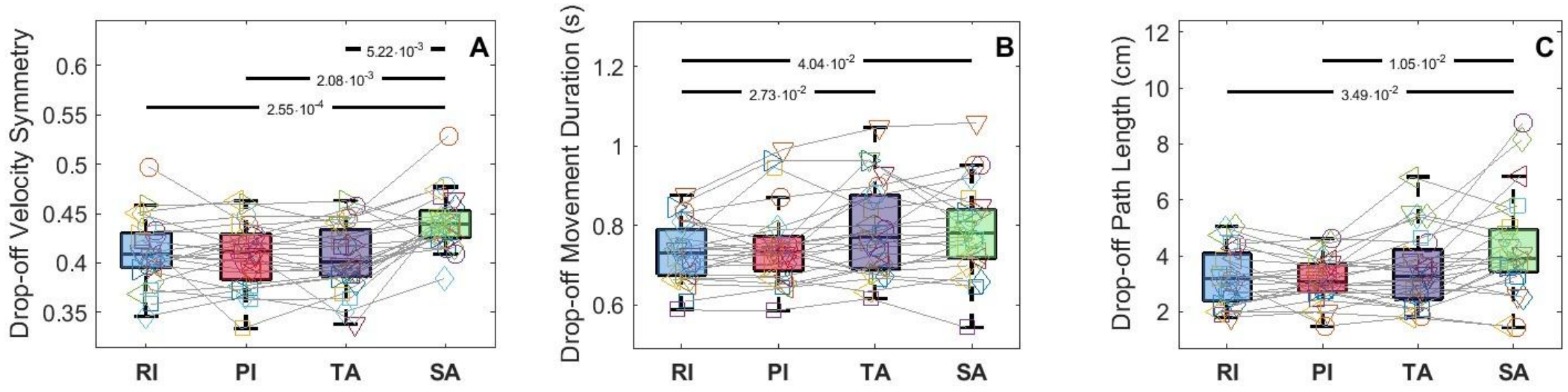

Figure 7. Effects of interaction algorithms on the drop-off movement. This figure shows the differences in the symmetry of the velocity profile (A, $\eta_p^2$ = 0.42), the movement duration for the drop-off movement (B, $\eta_p^2$ = 0.26), and peg path lengths for the drop-off movements (C, $\eta_p^2$ = 0.33) of the *robot-initiation* (RI), *participant-initiation* (PI), *temporal-alignment* (TA) and *spatiotemporal alignment* (SA) conditions. Significant differences are indicated by lines above the boxes, with the Šidák-corrected *p*-value embedded in the line.

Overall, the results indicate that *robot-initiation* offered significant benefits in terms of efficiency, closer pickup endpoints, and movement symmetry during pickup and drop-off phases. *Spatiotemporal-alignment* shared some temporal and spatial characteristics with *robot-initiation* but exhibits notable differences, including altered drop-off dynamics. These findings underscore the nuanced impacts of interaction algorithms on task performance, the implications of which will be discussed further in the General Discussion.

*Experiment 2: Effect of velocity profile of the robotic effector*

This experiment examined the effects of the robot-held peg velocity profile on participant movement performance. As in Experiment 1, differences between conditions are apparent in the raw data, which highlight key features of the experimental manipulations (see Figure 8). Notably, the constant-acceleration condition shows longer peg manipulation durations compared to the other profiles, while the biphasic-velocity condition exhibits less symmetrical velocity profiles—both trends mirroring some of the large effects reported below.

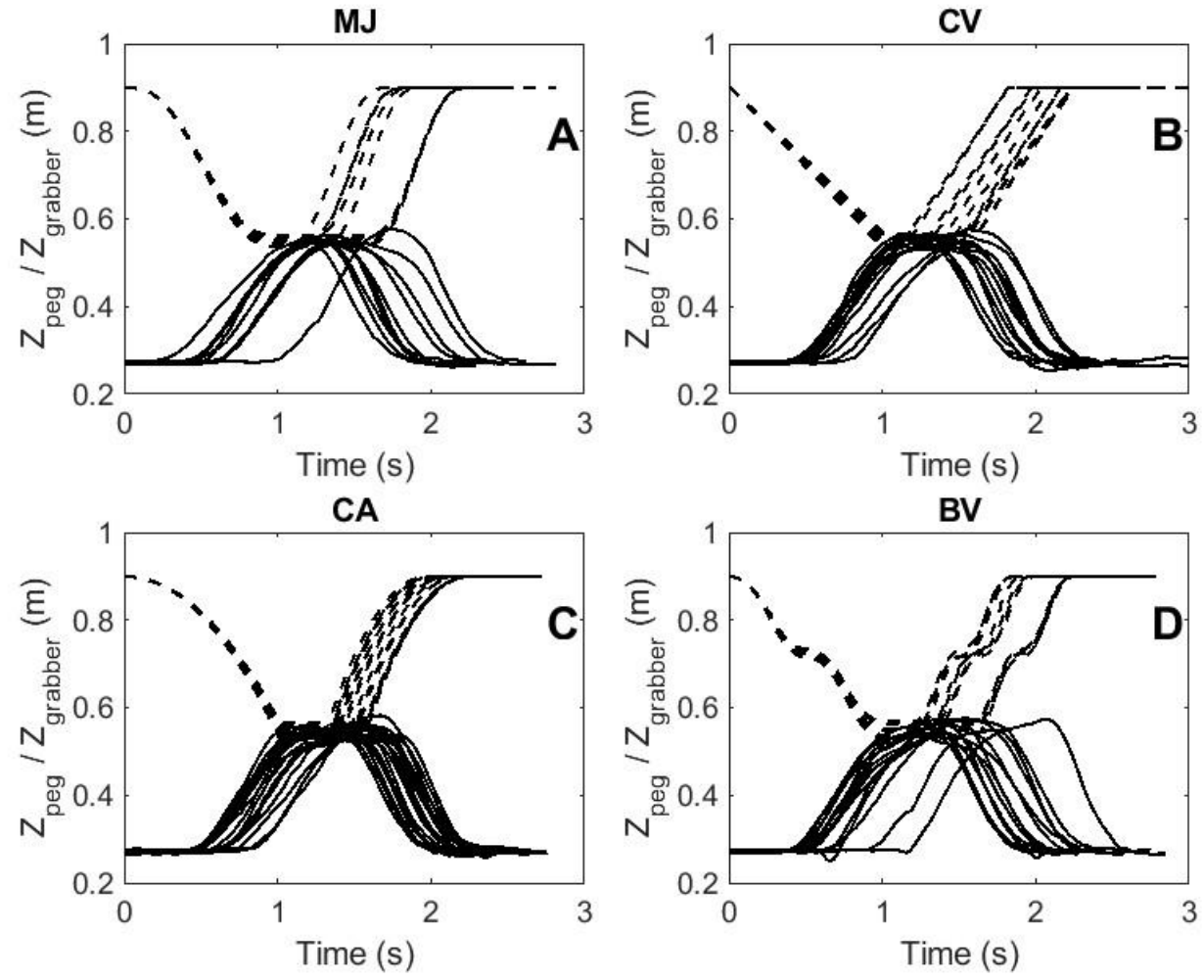

Figure 8. Raw Z-trajectories (i.e., unfiltered with no interpolation) of a participant, for the peg attached to the robot arm and the tip of participant-held grabber illustrating differences between *minimum-jerk* (MJ, panel A), *constant-velocity* (CV, panel B), *constant-acceleration* (CA, panel C), and *biphasic-velocity* (BV, panel D).

Several large effects (see Figure 9) suggested benefits of both *minimum-jerk* and *biphasic-velocity* trajectories for peg manipulation (i.e., the phase between the end of the participants forward movement and the moment the peg is successfully picked up), which resulted in shorter manipulation durations and path lengths. Moreover, vertical and depth pickup endpoints were also closer to the peg compared to *constant-acceleration* and *constant-velocity*, while lateral endpoints were closer compared to *constant-velocity* (for *biphasic-velocity* the lateral endpoint was also closer to the peg than for *constant-acceleration*).

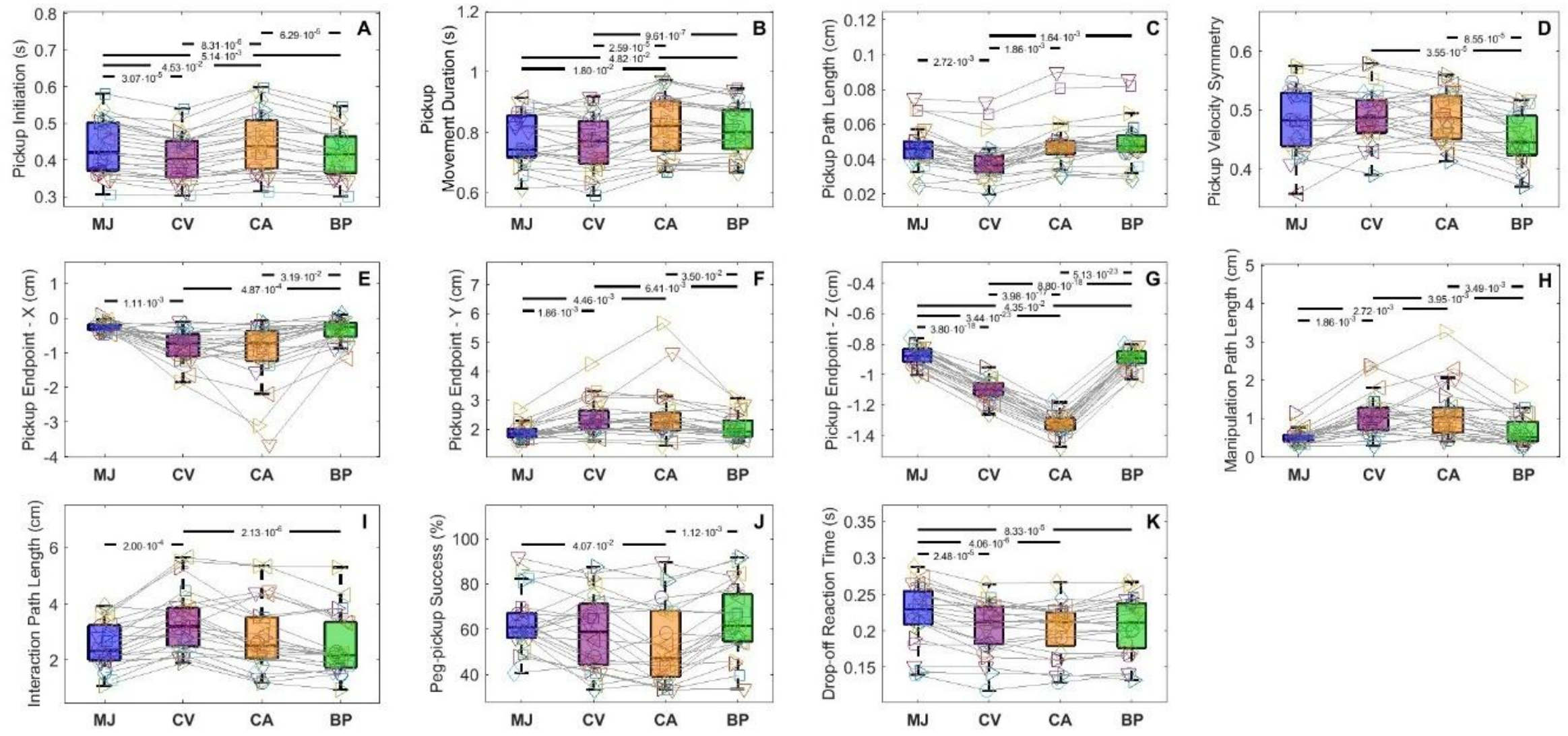

Figure 9. Large effects in Experiment 2 specific to the overall temporal performance and the pickup movement between the peg velocity profile conditions. Boxplots displaying the differences between conditions including pickup initiation (A, $\eta_p^2$ = 0.64), pickup movement duration (B, $\eta_p^2$ = 0.51), pickup path length (C, *W* = 0.56), pickup velocity symmetry (D, $\eta_p^2$ = 0.36), pickup endpoint- X (E, $\eta_p^2$ = 0.40), pickup endpoint-Y (F, *W* = 0.60), pickup endpoint-Z (G, $\eta_p^2$ = 0.99), manipulation path length (H, *W* = 0.60), interaction path length (I, $\eta_p^2$ = 0.42), peg-pickup success (J, $\eta_p^2$ = 0.30), and drop-off reaction time (K, $\eta_p^2$ = 0.59), between conditions *minimum-jerk* (MJ), *constant-velocity* (CV), *constant-acceleration* (CA), and *biphasic-velocity* (BV). Significant differences are indicated by lines above the significant conditions, with the Šidák-corrected *p*-value embedded in the line002E

*Minimum-jerk* and *biphasic-velocity* conditions further differed from *constant-acceleration*, with higher peg pickup success. Furthermore, movement initiation was earlier for *minimum-jerk, biphasic-velocity,* and *constant-velocity* conditions compared to the *constant-acceleration* condition. However, depth undershoots were smaller for *minimum-jerk* compared to *biphasic.* Interestingly, drop-off reaction times were later in the *minimum-jerk* condition compared to all other conditions. In contrast, *biphasic-velocity* profiles produced less symmetrical velocity profiles during pickup than both *constant-acceleration* and *constant-velocity.* Although pickup movement durations were shorter under *minimum-jerk* and *constant-velocity* than *biphasic-velocity* and *constant-acceleration*, pickup paths were straightest in the *constant-velocity* condition. *Constant-velocity* also yielded earlier pickup initiations, and smaller pickup depth undershoots compared to *constant-acceleration*, with longer manipulation phase durations for *constant-velocity* than for *constant-acceleration*.

*Experiment 3: Effect of timing of task-relevant information specification*

The goal of this experiment was to assess how the timing with which the required pickup orientation became visually specified influenced human performance during the robot-to-human handover task. A large effect of timing of peg rotation was found for pickup movement time, with longer pickup movements for *late-rotation* compared to all other conditions (Figure 10). Potentially as a result (due to the robot only keeping the peg in place for 550ms), a large effect for peg-pickup success showed that success was lower for *late-rotation* than for both *pretrial-rotation* and *synced-rotation*.

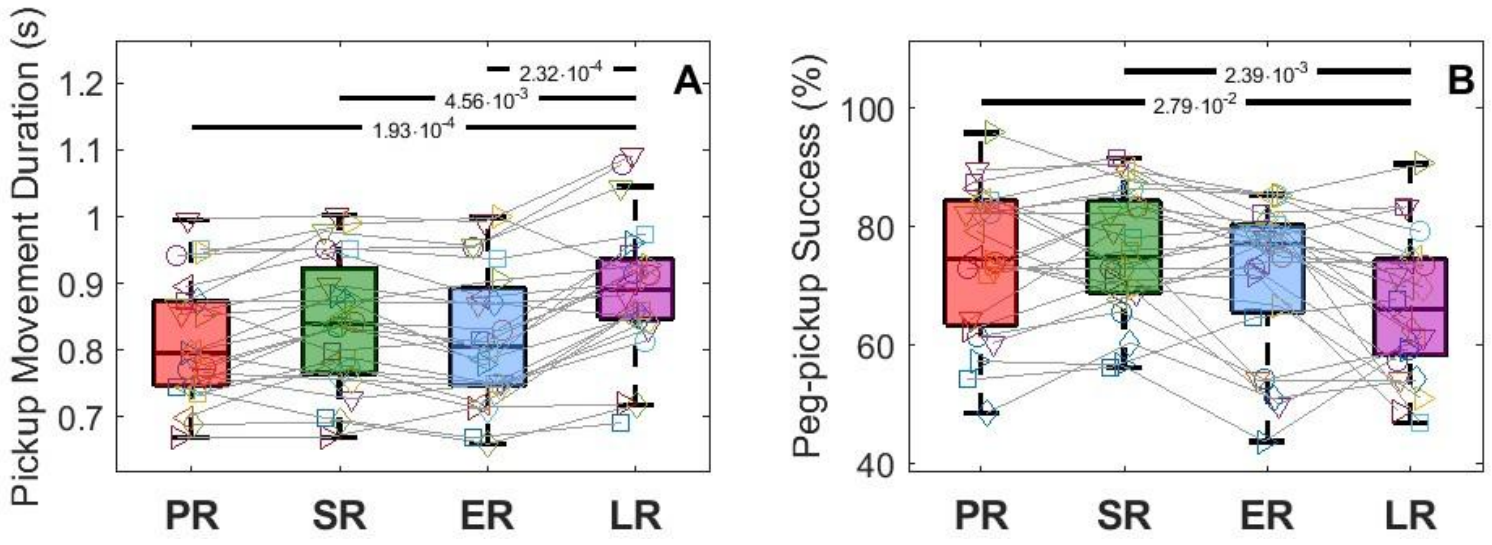

Figure 10: Large effects of peg rotation timing in Experiment 3. Boxplots displaying the differences between conditions including movement time for the pickup (A, $\eta_p^2$ = 0.48) and peg pickup success (B, $\eta_p^2$ = 0.29), between conditions pretrial-rotation (PR), synced rotation (SR), early-rotation (ER), and late rotation (LR). Significant differences are indicated by lines above the significant conditions, with the Šidák-corrected *p*-value embedded in the line.

*Summary of the findings*

Experiment 1 revealed benefits for *robot-initiation* compared to other interaction algorithms, resulting in faster pickup initiations, quicker and more efficient movements and straighter paths with more symmetrical velocity profiles. Although less pronounced than for *robot-initiation*, *spatiotemporal-alignment* also showed benefits compared to *participant-initiation* and *temporal alignment*, as evident from earlier initiations, shorter pickup durations, and more efficient (i.e., straighter) paths. Experiment 2 demonstrated strong benefits of *minimum-jerk* trajectories, including shorter movement durations, reduced displacement, and closer pickup endpoints. *Biphasic-velocity* profiles shared some of these benefits, with shorter movement durations and closer pickup endpoints compared to *constant-velocity* and *constant-acceleration* profiles. Finally, Experiment 3 found that *late-rotation* hindered pickup efficiency compared to all other conditions, in addition to decreasing pickup success rate compared to *synced-rotation* and *pretrial-rotation*.

**Discussion**

This study examined how different movement characteristics of a robotic effector influence human movements during robot-to-human handover tasks; a key focus was on how movement information provided by the robotic effector supports coordination of the human partner. We analyzed spatial and temporal aspects of movement performance during object handovers to quantify the effects of different modes of spatiotemporal control, the robotic effector's velocity profile, and the timing with which the required pickup orientation became specified in the visible motion of the robotic effector. As noted above, the analyses revealed several large effects; the Results section and following discussion focuses on the subset of these we judged most relevant and impactful; for completeness all large effects are reported in the Supplementary Information.

Experiment 1 examined manipulated two aspects related to how humans coordinate their movements in space and time. Firstly, we varied who initiated the interaction, which determined whether participants had access to early visual information about the future spatiotemporal evolution of the handover. Secondly, we included two other conditions in which the robotic effector adjusted to participant movements were included to examine the importance of who controls the spatiotemporal aspects of the handover. In these conditions the robotic effector always arrived at the interaction plane at the same time, or at the same place and time, as the participant-controlled grabber object, which effectively meant the participants controlled these aspects of the handover. Although the robot did not determine the exact moment of object acquisition, as it remained at the handover location until the participant picked up the peg, its arrival at the interaction zone did control the earliest opportunity for peg pick up. Thus, the conditions differed primarily in how they shaped the spatiotemporal evolution of the handover leading up to pick up. We compared these 4 conditions to establish how they impacted movement characteristics.

The key observation in Experiment 1 was that pickup movements were initiated earlier and executed more directly when the robotic effector initiated the interaction with a preplanned trajectory than when participants initiated the interaction and the robotic effector followed either a preplanned or temporally aligned strategy. Performance in the adaptive spatiotemporal-alignment condition was comparable to that observed in the robot-initiated condition, and generally exceeded that of temporal alignment alone, yielding earlier movement initiation and shorter movement paths. In addition, velocity profiles in the adaptive conditions exhibited longer deceleration phases, a characteristic commonly associated with increased reliance on feedback control [22, 42, 43]. A likely advantage of the robot-initiated condition was that participants could use the robotic effector's early movement to anticipate the forthcoming handover and begin organizing their own movement accordingly [20]. This information was not available in the participant-initiated conditions, suggesting that early movement information contributed substantially to the observed differences. At the same time, the similar performance observed in the robot-initiated and adaptive spatiotemporal-alignment conditions indicates that opportunities for online adjustment may partially compensate for the absence of such early information.

Although adaptive spatiotemporal alignment reduced the need for participants to continuously monitor the robotic effector in order to achieve spatiotemporal convergence, the prolonged deceleration phase observed in both adaptive conditions suggests that online visual feedback continued to play an important role during the final stages of the  movement [9, 19, 22]. Thus, rather than adopting a largely feedforward strategy, participants appeared to use the spatiotemporal certainty afforded by adaptation to support ongoing visual regulation of the pickup movement.

In both human-human interaction and HRI, the partner initiating movement is often considered to assume a leader role, whereas the responding partner assumes a follower role [37]. The present paradigms and results, however, suggest that movement initiation, information provision, and control over interaction dynamics can be dissociated. A participant may initiate movement while the robotic effector continues to determine critical spatiotemporal properties of the handover, whereas in adaptive conditions substantial control over those properties may be effectively transferred to the participant. Rather than by who simply moves first, leader-follower roles may be more appropriately characterized in terms of how information and control are distributed across the interaction. Beyond this, leader-follower distinctions are likely task-dependent [45, 47]; in the present task, the robotic effector retained responsibility for transporting and presenting the object, meaning that movement initiation did not fully determine control of the interaction.

Experiment 2 compared human movement performance for robotic effectors executing different preplanned velocity profiles. The underlying motivation for this manipulation was multifold. Firstly, comparing different velocity profiles, including human-like smooth profiles, can be informative with regard to human preferences [55], which may for instance be related to the human ability to perceive biological motion [7, 15, 28, 29, 38, 60]. Secondly, interceptive movements, including final positioning, are known to depend on the exact motion of the target object [9, 10, 21] in a way that could influence performance in our handover task, which can be considered a form of interception.

The main result of Experiment 2 was that the hand-held implement was positioned closer to the peg before the final pick up with more efficient manipulations for *minimal jerk* and *biphasic* trajectories. Only relative to *constant acceleration* trajectories condition did this translate to higher pickup success. The present findings are consistent with earlier reports that human-like smooth trajectories facilitate human-robot interactions [15, 29, 38, 56, 60]. One explanation is that humans possess extensive experience coordinating actions with such trajectories and may therefore extract task-relevant information from them more efficiently. While specialized mechanisms for processing biological motion may contribute to these effects [7, 28, 35], the present results do not require such an explanation because the manipulated variable primarily concerned end-effector motion rather than detailed joint kinematics. Moreover, because minimal jerk and biphasic trajectories both involve a smooth deceleration before reaching the interception zone, the present results suggest that the benefits of smooth trajectories may be concentrated in this final phase of the movement.

The aforementioned benefits can be understood from the interception literature. Visual information about the initial motion of the to-be-intercepted objects is known to shape initial interceptive movements [17] which are subsequently adapted to changes in target position and velocity through continuous monitoring of visual information, whenever available [19]. If early visual information is salient and accurate, interceptive movements tend to involve more efficient interception, because participants can guide actions adequately from initiation based on prior experience, with appropriate movement adjustments throughout the movement if needed, or both [9, 19]. Critically, targets decelerating before interception are generally easier to intercept [25, 52]; this can be understood from the smaller target displacement closer to interception (making the action more akin movements to stationary targets). This afforded picking up the peg while it was still in motion, albeit with very little displacement, which may be particularly beneficial for tasks that require late-stage spatial and temporal precision. When target displacement is concentrated within the later part of a trajectory, as seen with the *constant-acceleration* condition, this in-motion pick up is harder because the object is moving very fast; rather, participants needed more time both to start their movement and to pick up the peg, likely to allow for more visual feedback-based adjustments.

Experiment 3 examined the role of the timing with which the required pickup orientation became visually specified, which was manipulated by varying when the robotic effector rotated the peg during its forward movement. Later rotation resulted in slower forward movements (longer duration) and lower pickup performance compared to conditions in which rotation occurred earlier or synchronously with translation. Since the required pickup orientation only became visually specified late during the movement, participants could not fully organize their actions from the outset and online adjustments were therefore required, increasing movement duration. The later specification of the required pickup orientation also reduced pickup success. Together, these findings suggest that delaying the specification of action-relevant information can impair handover performance; by implication, earlier specification may facilitate coordination [20].

Some manipulations influenced behaviour beyond the functionally interactive phase of the task, affecting the subsequent transport of the peg to the receptacle. In Experiment 1, for example, several conditions differed in both the pickup and drop-off phases. These effects may reflect differences in the state from which the transport movement was initiated, persistence of movement organization established during the pickup phase, or both. Regardless of the underlying mechanism, the findings suggest that the consequences of human-robot coordination can extend beyond the handover itself. Optimizing collaborative interactions may therefore require consideration of the entire task sequence, as changes in the efficiency of sequential movements may influence overall task performance directly through movement duration and indirectly through cumulative factors such as physical effort and fatigue [Peternel et al., 2019].

*Implications and future directions*

Collectively, these findings suggest that robotic effectors can facilitate coordination by providing movement information that is provided earlier, is clearer, or is easier for humans to use during action. By implication, rather than increasing system complexity through additional communication mechanisms, robotic effectors may facilitate coordination by specifying task-relevant action requirements earlier and more clearly through their movement. Beyond implicit communication through kinematics (i.e., legibility [15, 20]), the detailed kinematics themselves can also impact the quality of the interaction; particularly smooth deceleration close to the interaction appears beneficial, at least in the kind of interaction studied here.

Virtual reality afforded a particular class of experiments involving systematic manipulation of movement characteristics that would be difficult to implement consistently and safely using physical robotic systems, while retaining key characteristics of handover behaviour. Although the present results provide no clear indication that the use of VR fundamentally altered the coordination phenomena under investigation, further studies are needed to establish the extent to which they generalize to physical human-robot interactions. Movement behaviour [2, 51] and perception [3] in VR can differ from those observed in physical environments, partly due to factors such as altered depth perception [14] and reduced multisensory feedback [13]. The latter is particularly relevant given the absence of haptic sensations associated with physically interacting with a robotic partner, which may have increased participants' reliance on visual information [23] and thereby influenced both movement execution and task realism [18]. Whether such sensory features principally affect the physical aspects of the interaction in the present study, or also influence earlier stages involving movement initiation and coordination, remains an important question for future research. Addressing this issue will be essential for establishing the ecological validity of VR-based studies of human-robot interaction.

The work presented in this study provides several clear directions for future investigations. First of all, the exploratory nature of the study means the interpretations need future confirmation, either through explicit replication or more dedicated tests of the proposals. Second, although Experiment 1 suggested that both early specification of movement information and opportunities for adjustment may benefit performance, the present design did not fully dissociate these factors. It would therefore be interesting to consider a combination of these two main manipulations: having the robotic effector initiate the interaction with a movement in a specific direction, with adaptive updates starting only once the participant initiates their movement. The robotic effector's initial movement could provide information about the likely evolution of the interaction, allowing participants to begin organizing their movement before subsequently influencing the timing and location of the handover through online adjustments.

Future investigations should also examine the potential role of learning. The relatively limited exposure to each condition in the present study may have constrained participants' ability to adapt to the different interaction modes. This may have been particularly relevant for the adaptive conditions, for which details of the interaction depended intimately on participants' own movements and for which increased experience could allow more effective use of opportunities for online adjustments. In addition, the adaptive spatiotemporal-alignment condition was always administered last and, unlike the other conditions, allowed participants to influence the final handover location. Although neither issue affects the primary conclusions of the present study, future work should employ increased exposure, counterbalanced presentation orders, and more tightly controlled handover locations to determine the extent to which the observed advantages reflect learning, differences in handover geometry, the interaction algorithm itself, or interactions between these factors.

In Experiment 2 we contrasted smooth, human-like robot velocity profiles with other profiles, finding some benefits of smooth movements consistent with previous work [15, 29]. The results showed indications, however, that it is not just 'smoothness' of the trajectory that matters, but that the benefits of smooth trajectories may be concentrated in the final deceleration phase preceding object acquisition. Future studies should determine whether the observed effects arise from human-like movement characteristics more generally or specifically from movement conditions during the final approach to pick-up.

Finally, this research focused primarily on a relatively straightforward robot-to-human handover task. It remains to be determined whether the observed effects generalize to tasks involving different sensorimotor requirements, including handovers that require more tightly synchronized actions or more complex collaborative manipulations. Even within the present task, interceptive demands could be systematically increased by reducing the time for which the robotic effector remains at the handover location before departing. Manipulations such as these may help determine how the importance of movement kinematics and the timing with which task-relevant information becomes specified depends on the coordination demands of the task. More complex interactions may also engage psychological factors such as trust and acceptance of robotic partners [31], aspects unlikely to have strongly influenced behaviour in the present study.

**Conclusion**

This study examined how movement characteristics of a robotic effector influence human motor performance during robot-to-human handover tasks in virtual reality. Across the experiments, performance was influenced by how movement information was distributed between interaction partners, how clearly task-relevant action requirements were specified, and how movement kinematics shaped opportunities for coordination. In particular, the findings suggest that earlier specification of action-relevant information facilitates handover performance, while smooth trajectories, especially those involving a gradual deceleration before object acquisition, support more efficient interaction. The results also highlight that movement initiation, information provision, and control over interaction dynamics are related but distinct aspects of collaborative action. Together, these findings contribute to a better understanding of how humans coordinate their movements with robotic partners and suggest that robotic effectors may facilitate interaction by making task-relevant information more clearly specified within their movement. Future work should determine the extent to which these principles generalize to physical robotic systems and more complex forms of human-robot collaboration.

This research was funded through a postgraduate stipend from the Department for Economy (DfE Northern Ireland).

**Movement Information from a Robotic Effector Influences Human Performance During Object Handovers**

Róisín Keenan, Joost C. Dessing

School of Psychology, Queen’s University Belfast

**Supplementary Information**

**Contains:**

- **Supplementary Results – Experiments 1-3**
- **Supplementary Experiment (Experiment 4) - Methods and Results**
- **Supplementary Tables**
- **Supplementary References**

## Supplementary Results – Experiments 1-3

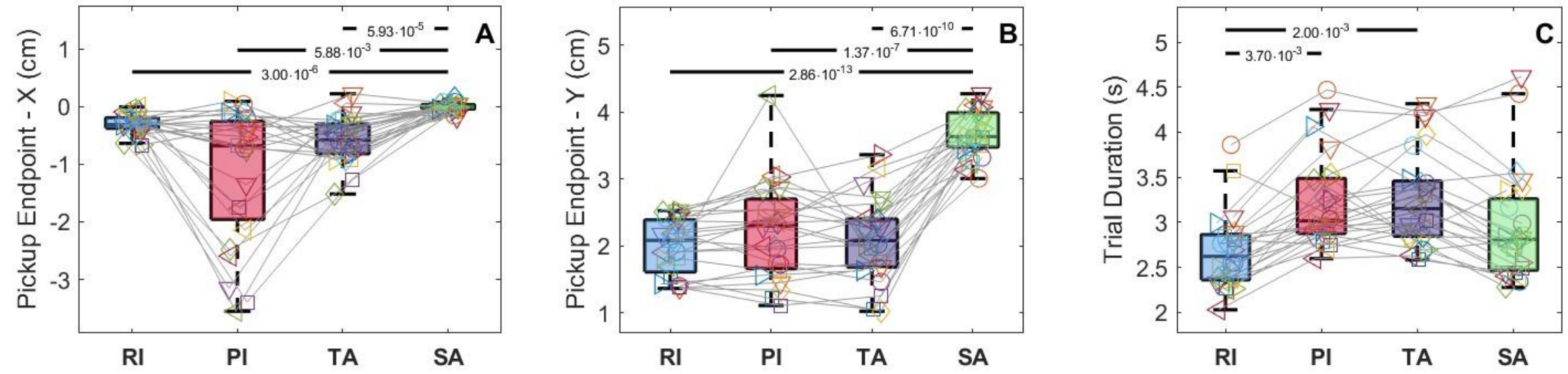

Figure S1: Additional large effects of interaction algorithms in Experiment 1 on the sideward (A) and upward position (B) of the end of the pickup movement, relative to the position of the top of the peg, and the trial duration (C). This figure shows the effects for the drop-off movements (C) of the robot-initiation (RI), participant-initiation (PI), temporal alignment (TA) and spatiotemporal alignment (SA) conditions. Significant differences are indicated by lines above the boxes, with the Šidák-corrected *p*-value embedded in the line. See main text and above for more details.

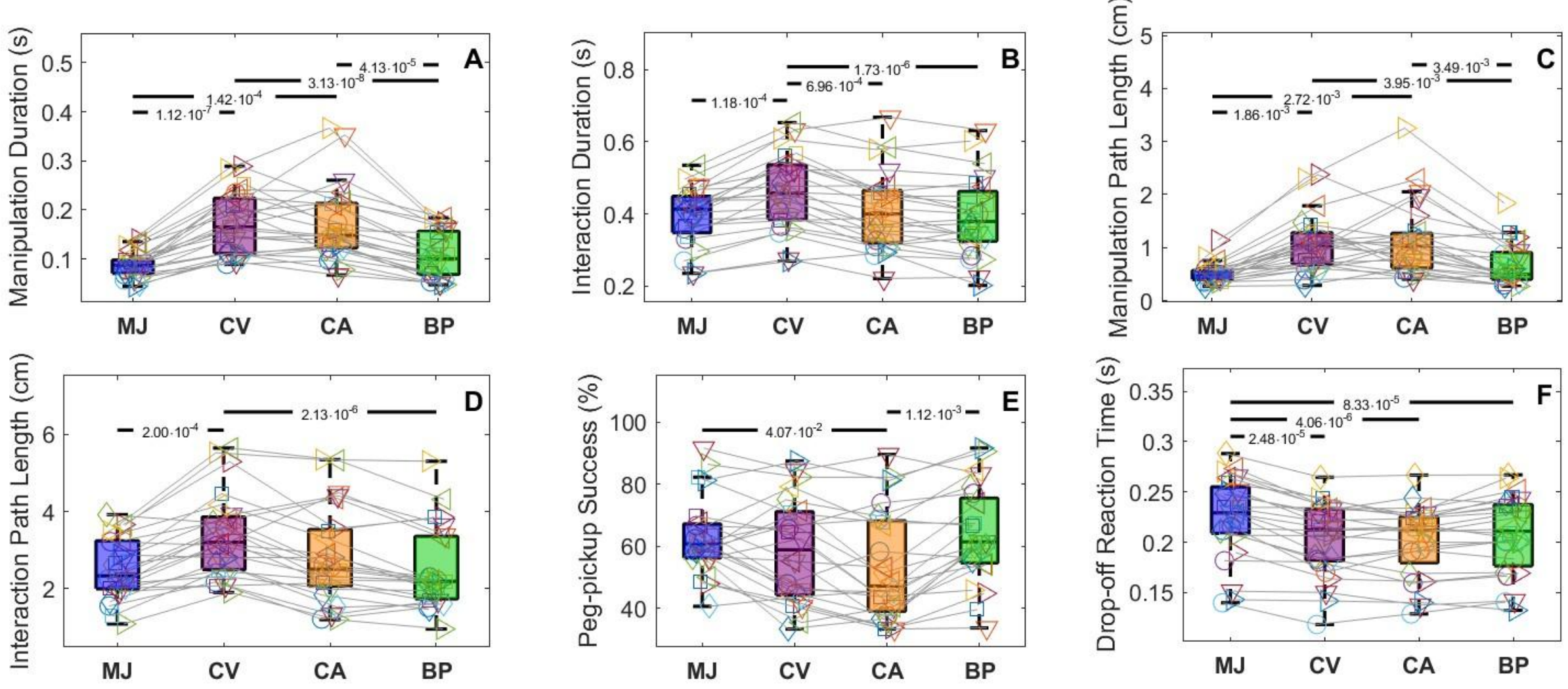

Figure S2: Additional large effects of robot velocity profile in Experiment 2 on the Manipulation Duration (A), Interaction Duration (B) , Manipulation Path Length (C), Interaction Path Length (D), Peg-pickup Success (E), and Drop-off Reaction Time (F). This figure shows the effects for the Minimal Jerk movements (MJ), Constant Velocity movements (CV), Constant Acceleration movements (CA), and Biphasic movements (BP). Significant differences

are indicated by lines above the boxes, with the Šidák-corrected *p*-value embedded in the line. See main text and above for more details.

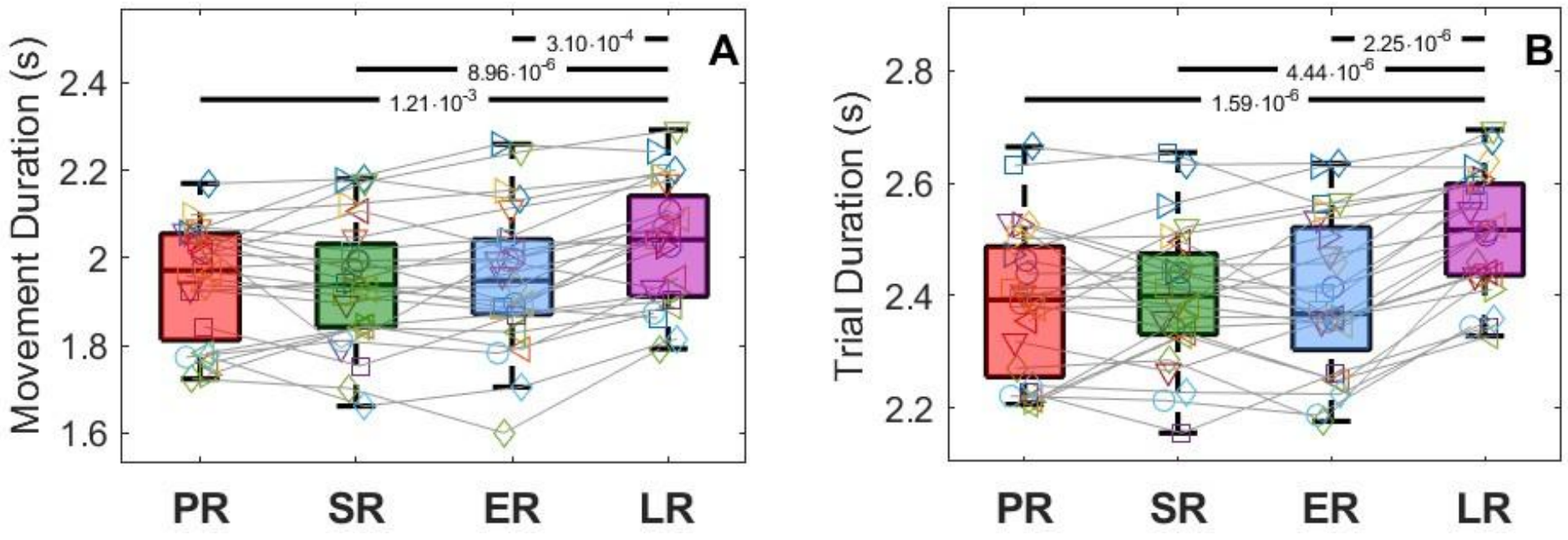

Figure S3: Additional large effects of peg rotation timing in Experiment 3 on Movement Duration (A) and Trial Duration (B) for conditions pretrial-rotation (PR), synced rotation (SR), early-rotation (ER), and late rotation (LR). Significant differences are indicated by lines above the significant conditions, with the Šidák-corrected p-value embedded in the line. See main text and above for more details.

**Supplementary Experiment – Experiment 4**

An additional experiment examined the role of the type of interaction partner and the visual salience of its joint rotations. Visual appearance of partners including robots in interactive task have been shown to affect performance [38, 65]. If joint kinematics matter for human performance in the robot-to-human handover task, visual salience of joint rotations could be expected to influence performance. Participants engaged with a human avatar or robot arm handing over the peg, with matching end-effector motion (based on offline recordings). We manipulated the elevation of the elbow of this partner in an attempt to manipulate visual salience of elbow rotation. The specific methods and results of this experiment are presented below, although otherwise this experiment involved the same general methods as Experiment 1 in the main text. In the text below, this experiment is referred to as Experiment 4, although it was conducted in combination with Experiment 1.

*Participants*

One participant who participated in Experiment 1 did not complete Experiment 4 (and so 1 new participant was recruited) - five were excluded due to technical issues involving peg movement inconsistencies between the virtual partners (see below), resulting in a final sample of 19 participants.

*Experimental setup*

This experiment aimed to explore whether the visual appearance of the task partner (i.e., *human* avatar/partner vs *robot* arm) influences the benefits of smooth human-like motion patterns, and to investigate how the visibility of joint motion of a robot influenced human interaction during a joint handover task. The end-effectors of the human avatar and robot arm (i.e., the peg) followed identical trajectories, as derived from three-dimensional (3D) motion tracking of a human actor. The trajectory kinematics were generated using previously recorded full-body motion tracking with the Xsens Awinda system (Movella Technologies B.V., Enschede, The Netherlands). The actor held their arm either mostly vertical (elbow down, 0° abduction; $E_{low}$), half raised (arm abducted about 45°; $E_{mid}$), or mostly horizontal (arm abducted around 90°; $E_{high}$). The actor completed 12 movements 350mm forward (on average) to four predefined target points (on a 30cm x 30cm square [touched with a forward-facing fist during the motion recording sessions]) presented on a vertical plank. Analyses of the reconstructed trajectory end points showed these were not always 30cm apart, with deviations existing in the z- and y-axis (< 6cm), and more prominently the x-axis (< 19cm). These trajectories were still included in the experimental design to add variance for the participant; critically, the trajectory end-point was not a variable of interest for the analysis and the same trajectories were used for both the human avatar and robot arm. We did exclude 3 trajectories based on visual inspection, which showed positional deviations that were likely reflective of tracking errors (see Figure S4).

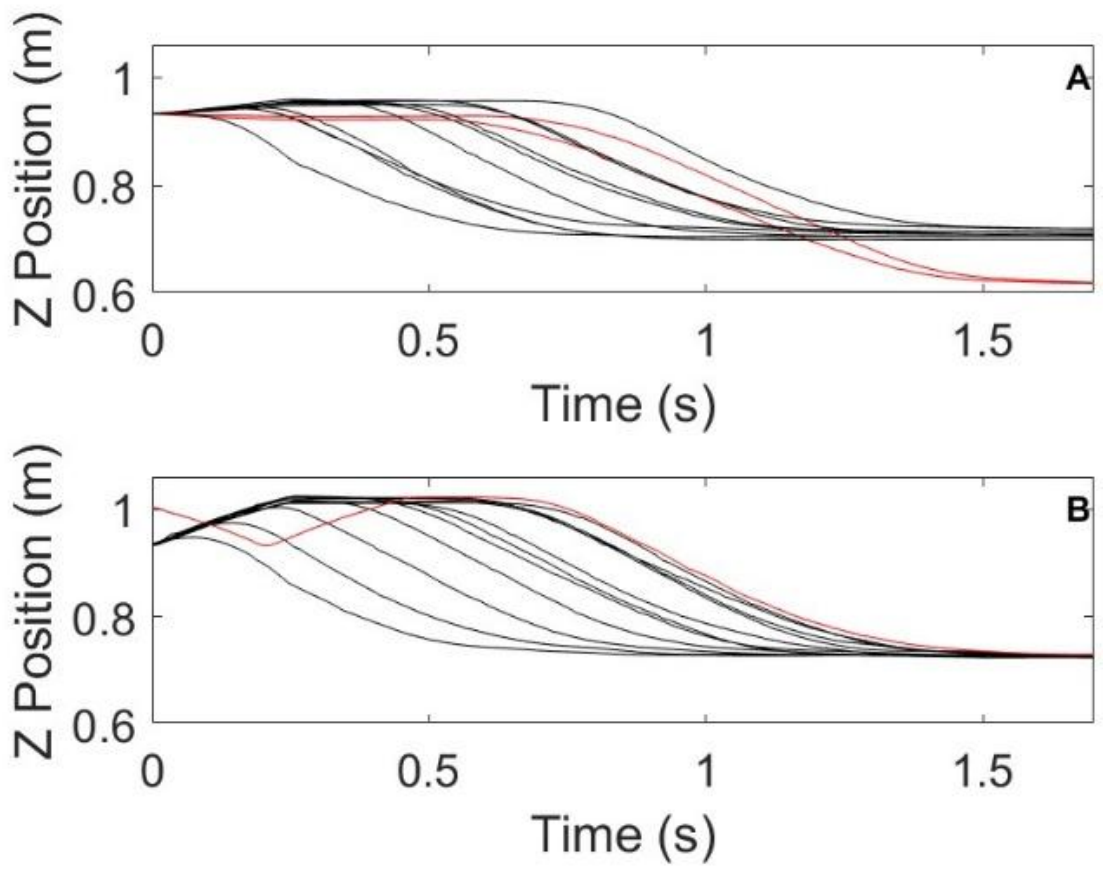

Figure S4: Previously recorded trajectories using the Xsens Awinda System. The top plot demonstrates the z-trajectories for the 45° elbow elevation and lower right final position with 2 excluded trajectories (based on position deviations) displayed in red. The bottom plot demonstrates the z-trajectories for the 90° elbow elevation and lower right final position with 1 excluded trajectory (based on initial movement) in red.

Besides the robot arm used in the other experiments, this experiment also involved a *human* partner, represented by a commercially available avatar Unity asset[2], configured to sit on a virtual chair at the virtual table across from the participant. The human avatar held a bar that had the peg base and peg attached (see

[2] Original asset available from the Unity Asset Store: *Real Human Player 02*, accessed 9 November 2022. https://assetstore.unity.com/packages/3d/characters/humanoids/humans/real-human-player-02162313.

Figure S5). Of the recorded animations, we only used movements of the right arm (i.e., we fixed the avatar shoulder in space in the virtual environment). We did add some minor motion animations to the avatar's left arm and head (such as slight left arm sway when moving the opposing arm) to prevent it from appearing unnatural (i.e., too stationary).

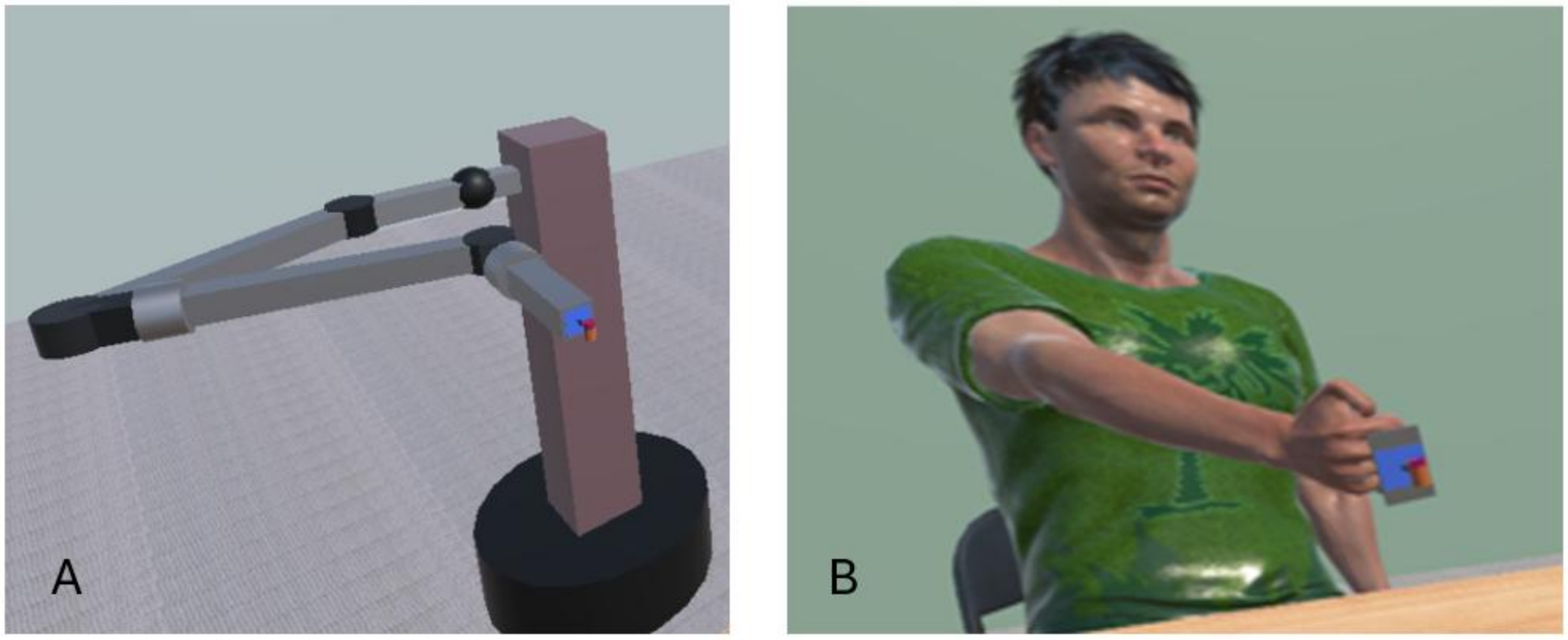

Figure S5: Virtual task partners. Panel A shows the virtual robot partner set-up, panel B shows the human partner implemented while using the same experimental framework. Both partners present the participants with a peg to pick up.

The experiment further explored how joint motion visibility affected performance in the collaborative manual handover task. It was expected that elevating the elbow position (adding abduction to the arm) may increase the salience of the joint kinematic information and thus increase the potential (predictive) benefits of biological motion perception. This experiment involved arm abductions of 0° (arm in a vertical plane), 45° (arm in a tilted plane), and 90° (arm in a horizontal plane, at shoulder height) (see Figure S6). Increased salience of the movement kinematics (through elevation of the elbow joint angle to a 90° abduction) was expected to aid performance.

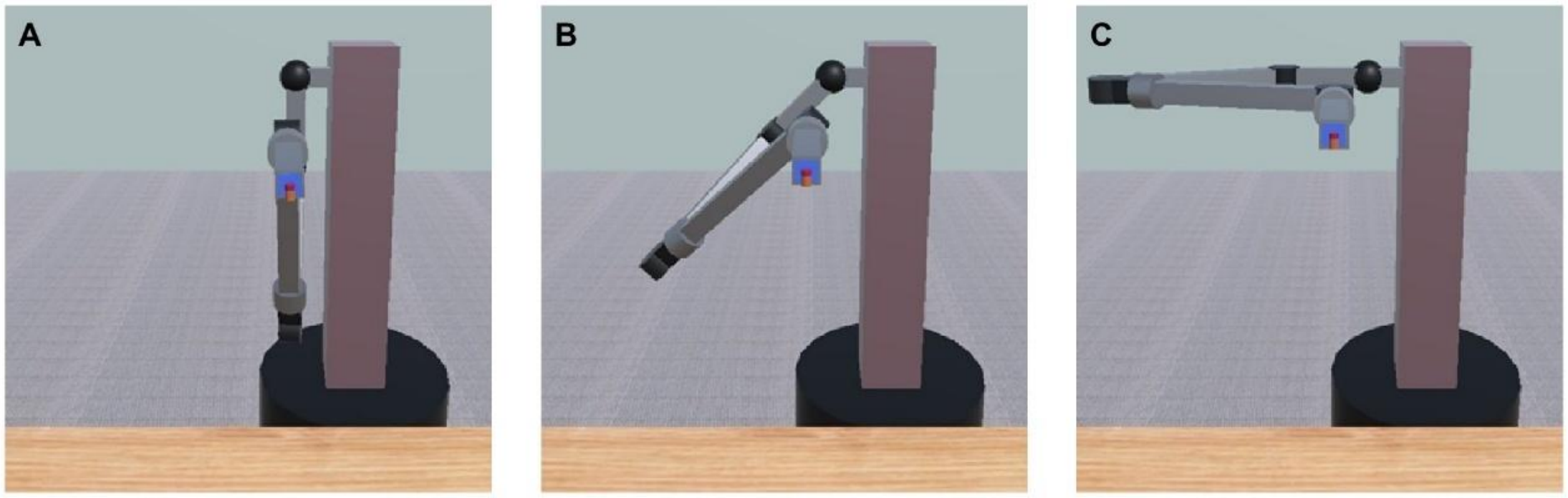

Figure S6: Robot 'elbow' elevations. 'Elbow' elevation was manipulated by abducting the robot arm at the 'shoulder' (the black sphere), rotating it around the z-axis by 0°(A), 45°(B), and 90°(C) to implement (conditions $E_{low}$, $E_{mid}$, and $E_{high}$, respectively).

*Procedures*

For each partner condition in the Experiment 4, participants completed 24 repetitions for all 12 conditions (final positions and elbow elevations). The 24 repetitions consisted of two repetitions of all available unique movements per condition. For those conditions with fewer than 12 available movements (due to the aforementioned exclusions), additional repetitions were randomly selected for each participant and partner condition to achieve the 24 repetitions (these were randomly assigned for both task partners). The total number of trials thus was 576 (288 per task partner). Trials were presented in blocks defined by the elbow elevation manipulations ($E_{low}$, $E_{mid}$, $E_{high}$), with the peg end position randomized within blocks. Five practice trials were completed at the start of each block - these were not included in the analyses. This experiment took around 60 minutes to complete.

*Data analyses*

This experiment involved mostly the same analyses as Experiment 1 in the main paper, but some experiment-specific analyses are described here. Data analyses revealed a number of trials where the peg trajectory – in particular its movement time - deviated from the planned trajectory. For this reason, trials were excluded if their trajectories exceeded 3% of the raw trajectory movement time (i.e., what the trajectory timing should have been). Participants for whom this exclusion criterion was met for more than 25% of their total trials per condition (i.e. > 24/96 trials) were excluded from the analyses, this led to the exclusion of 5 participants. Beyond this, visually detected signal deviations led to the exclusion of a further 17 trials from Experiment 4.

Normality tests were based on Shapiro-Wilk tests and using a cutoff of $\alpha < 0.05/N$ (where N = 6 for Experiment 4)). For post-hoc testing, the corrections for multiple comparisons depended on the factor: for the *task partner* conditions, $k = 2$ was used for 2 large effects with 1 comparison; for *joint visibility*, $k = 6$ was used based on 2 large effects each followed by 3 post hoc comparisons. See main text for full details on these procedures.

**Supplementary Results – Experiment 4: Human vs robot as virtual task partner**

In Experiment 4 we observed large effects of task partner for duration and path length of the pickup movement (Figure S7A, B), both being longer when interacting with the *robot* partner compared to the *human* partner. The experiment further explored how joint visibility (manipulated by variations in elevation of the robot elbow) affected performance in the collaborative handover task (Figure S7C, D). There was a large effect on vertical pickup endpoint, with $E_{low}$ resulting in a higher pick-up end-position than $E_{high}$, but differences did not pass our (stringent) significance criterion. Symmetry of the drop-off velocity profile was lower (less symmetrical) for the low elevation compared to both the mid and high elevation, with the latter having the most symmetrical drop-off velocity profile.

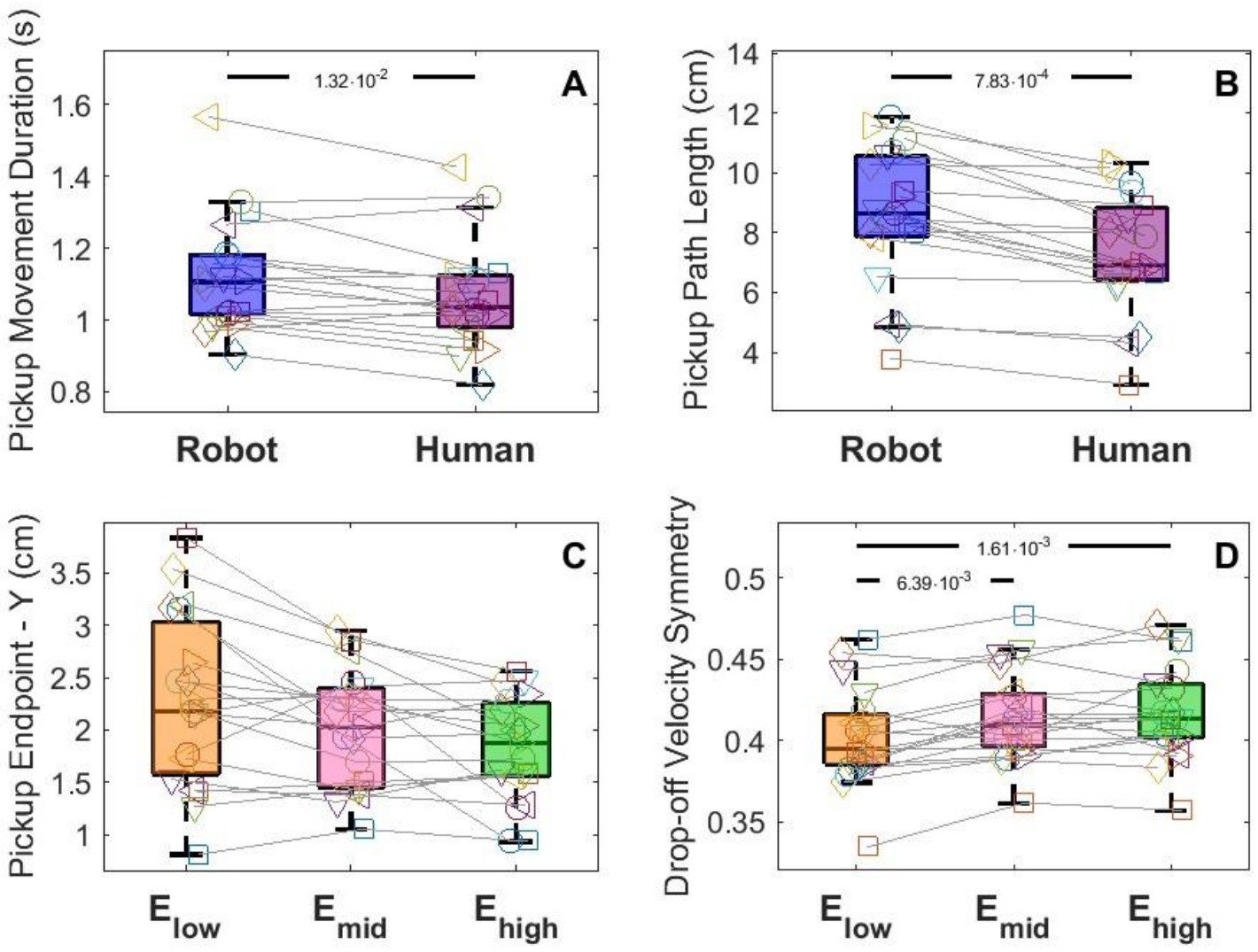

Figure S7: Large effects in Experiment 2 of task partner and elbow elevation. Illustrating differences of task partner on pickup movement time (A) and path length (B) and effects of elbow elevation levels on vertical error (C) and the velocity symmetry of the drop-off movement (D). Significant differences are indicated by lines above the boxes, with the Šidák-corrected *p*-value embedded in the line.

*Summary*

This experiment showed that interacting with a *human* partner led to shorter pickup durations with reduced displacement (straighter path lengths), in addition to salient visual kinematics resulting in closer vertical endpoints and resulting in more symmetrical velocity profiles of drop-off movements.

**Supplementary Tables**

**Table S1: Statistics for variables showing a large effect for interaction algorithm (Experiment 1)**

| Variable | Sphericity Method | Main effect of interaction algorithm - statistics |
|---|---|---|
| Vertical Pickup Endpoint (Y) | GG | $F(2.25, 51.65) = 89.51$, $p = 2.43{\cdot}10^{-18}$, $\eta_p^2 = 0.80$ |
| Pickup Movement Time | HF | $F(2.52, 57.85) = 29.68$, $p = 2.73{\cdot}10^{-11}$, $\eta_p^2 = 0.58$ |
| Pickup Path Length | SA | $F(3, 69) = 26.02$, $p = 2.27{\cdot}10^{-11}$, $\eta_p^2 = 0.53$ |
| Pickup Initiation | GG | $F(2.19, 50.39) = 25.68$, $p = 8.42{\cdot}10^{-9}$, $\eta_p^2 = 0.53$ |
| Pickup Velocity Symmetry | SA | $F(3, 69) = 18.86$, $p = 9.84{\cdot}10^{-9}$, $\eta_p^2 = 0.45$ |
| Depth Pickup Endpoint (Z) | SA | $F(3, 69) = 16.71$, $p = 2.76{\cdot}10^{-7}$, $\eta_p^2 = 0.42$ |
| Drop-off Velocity Symmetry | GG | $F(2.15, 49.32) = 16.49$, $p = 2{\cdot}10^{-6}$, $\eta_p^2 = 0.42$ |
| Lateral Pickup Endpoint (X) | GG | $F(1.22, 28.02) = 14.94$, $p = 3.00{\cdot}10^{-4}$, $\eta_p^2 = 0.39$ |
| Drop-off Path Length | GG | $F(2.10, 48.19) = 11.31$, $p = 7.45{\cdot}10^{-5}$, $\eta_p^2 = 0.33$ |
| Drop-off Movement Time | SA | $F(3, 69) = 7.90$, $p = 1.33{\cdot}10^{-4}$, $\eta_p^2 = 0.26$ |
| Overall Trial Duration | N/A | $\chi2(3) = 36.05$, $p = 7.31{\cdot}10^{-8}$, $W = 0.50$ |

**N.B ordered with parametric variables first and based on effects size (high to low). SA = Sphericity Assumed; GG = Greenhouse-Geisser; HF = Huynh Feldt; Dfs = Degrees of Freedom; $\eta_p^2$ = Partial-Eta Squared.**

**Table S2: Statistics for variables showing a large effect for robot velocity profile (Experiment 2)**

| Variable | Sphericity Method | Main effect of peg velocity profile - statistics |
|---|---|---|
| Depth Pickup Endpoint (Z) | HF | $F(2.53, 58.13) = 1559.10$, $p = 1.06 \cdot 10^{-53}$, $\eta_p^2 = 0.99$ |
| Manipulation Duration | GG | $F(1.91, 43.84) = 44.24$, $p = 5.24 \cdot 10^{-11}$, $\eta_p^2 = 0.66$ |
| Pickup Initiation | GG | $F(1.85, 42.59) = 41.38$, $p = 2.37 \cdot 10^{-10}$, $\eta_p^2 = 0.64$ |
| Drop-off Reaction | SA | $F(3, 69) = 32.48$, $p = 3.32 \cdot 10^{-13}$, $\eta_p^2 = 0.59$ |
| Trial Duration | GG | $F(2.09, 48.03) = 29.57$, $p = 2.75 \cdot 10^{-9}$, $\eta_p^2 = 0.56$ |
| Pickup Movement Time | GG | $F(2.21, 50.73) = 23.92$, $p = 1.93 \cdot 10^{-8}$, $\eta_p^2 = 0.51$ |
| Interaction Duration | GG | $F(2.20, 50.62) = 20.70$, $p = 1.20 \cdot 10^{-7}$, $\eta_p^2 = 0.47$ |
| Interaction Path Length | GG | $F(2.06, 47.29) = 16.83$, $p = 2.49 \cdot 10^{-6}$, $\eta_p^2 = 0.42$ |
| Total Movement Time | SA | $F(3, 69) = 16.38$, $p = 2.93 \cdot 10^{-9}$, $\eta_p^2 = 0.42$ |
| Lateral Pickup Endpoint (X) | GG | $F(1.45, 33.37) = 15.01$, $p = 1.06 \cdot 10^{-4}$, $\eta_p^2 = 0.40$ |
| Pickup Velocity Symmetry | GG | $F(2.02, 46.55) = 12.17$, $p = 3.70 \cdot 10^{-5}$, $\eta_p^2 = 0.36$ |
| Peg-Pickup Success | SA | $F(3, 69) = 9.83$, $p = 1.77 \cdot 10^{-5}$, $\eta_p^2 = 0.30$ |
| Manipulation Path Length | N/A | $\chi 2(3) = 46.85$, $p = 3.74 \cdot 10^{-10}$, $W = 0.65$ |
| Vertical Pickup Endpoint (Y) | N/A | $\chi 2(3) = 42.95$, $p = 2.52 \cdot 10^{-9}$, $W = 0.60$ |
| Pickup Path Length | N/A | $\chi 2(3) = 40.60$, $p = 7.95 \cdot 10^{-9}$, $W = 0.56$ |

**N.B ordered with parametric variables first and based on effects size (high to low). SA = Sphericity Assumed; GG = Greenhouse-Geisser; HF = Huynh Feldt; Dfs = Degrees of Freedom; $\eta_p^2$ = Partial-Eta Squared.**

**Table S3: Statistics for variables showing a large effect for rotation timing (Experiment 3)**

| Variable | Sphericity Method | Main effect of Rotation Timing – Statistics |
|---|---|---|
| Trial Duration | SA | $F(3, 69) = 26.79$, $p = 1.33 \cdot 10^{-11}$, $\eta_p^2 = 0.54$ |
| Pickup Movement Time | HF | $F(2.51, 57.69) = 21.18$, $p = 1.45 \cdot 10^{-8}$, $\eta_p^2 = 0.48$ |
| Total Movement Time | SA | $F(3, 69) = 14.63$, $p = 1.79 \cdot 10^{-7}$, $\eta_p^2 = 0.39$ |
| Peg-Pickup Success | SA | $F(3, 69) = 9.42$, $p = 2.69 \cdot 10^{-5}$, $\eta_p^2 = 0.29$ |

**N.B ordered from high – low effect size. SA = Sphericity Assumed; GG = Greenhouse-Geisser; HF = Huynh Feldt; Dfs = Degrees of Freedom; $\eta_p^2$ = Partial-Eta Squared.**

**Table S4: Statistics for variables showing a large effect for task partner ([Supplementary] Experiment 4)**

| Variable | Sphericity Method | Main effect of task partner - statistics |
|---|---|---|
| Pickup Path Length | HF | $F(1, 18) = 34.46$, $p = 1.47 \cdot 10^{-5}$, $\eta_p^2 = 0.66$ |
| Pickup Movement Duration | HF | $F(1, 18) = 11.45$, $p = 3.31 \cdot 10^{-3}$, $\eta_p^2 = 0.39$ |

**N.B ordered from high – low effect size. SA = Sphericity Assumed; GG = Greenhouse-Geisser; HF = Huynh Feldt; Dfs = Degrees of Freedom; $\eta_p^2$ = Partial-Eta Squared.**

**Table S5: Statistics for variables showing a large effect for joint visibility ([Supplementary] Experiment 4)**

| Variable | Sphericity Method | Main effect of joint visibility - statistics |
|---|---|---|
| Drop-off Velocity Symmetry | SA | $F(2, 36) = 12.24$, $p = 8.78 \cdot 10^{-5}$, $\eta_p^2 = 0.41$ |
| Vertical Pickup Endpoint (Y) | HF | $F(1.62, 29.14) = 6.89$, $p = .006$, $\eta_p^2 = 0.28$ |

**N.B ordered from high – low effect size. SA = Sphericity Assumed; GG = Greenhouse-Geisser; HF = Huynh Feldt; Dfs = Degrees of Freedom; $\eta_p^2$ = Partial-Eta Squared.**

**Supplementary references**